\documentclass[]{osunlp}

\usepackage{amsmath}
\usepackage{amssymb}
\usepackage{mathtools}
\usepackage{bm}
\usepackage{algorithm}
\usepackage{algpseudocode}
\usepackage{paralist}
\usepackage[most]{tcolorbox}
\usepackage{listings}
\usepackage[dvipsnames,table,xcdraw]{xcolor}
\usepackage{booktabs}
\usepackage{array}
\usepackage{tabularx}
\usepackage{ragged2e}
\usepackage{makecell}
\usepackage{multirow}
\usepackage{tikz}
\usepackage{pgfplots}
\usepackage[edges]{forest}
\usepackage{wrapfig}
\usetikzlibrary{pgfplots.groupplots, matrix}
\pgfplotsset{compat=1.18}
\usepackage{enumitem}
\usepackage{comment}
\usepackage{graphicx}
\usepackage{multirow}
\usepackage{stackengine}
\usepackage{colortbl} %
\usepackage{anyfontsize}

\definecolor{purple}{HTML}{8B1E1E}
\definecolor{navyblue}{HTML}{BB0000}
\definecolor{citecolor}{HTML}{BB0000}
\definecolor{lightgray}{gray}{0.9}
\definecolor{blanchedalmond}{rgb}{1.0, 0.92, 0.8}
\definecolor{cerise}{rgb}{0.871, 0.192, 0.388}

\definecolor{TaskBG}{HTML}{FAEAEA}        %
\definecolor{StateBG}{HTML}{F5F5F7}       %
\definecolor{ExpertBG}{HTML}{EAF4E3}      %
\definecolor{IWMBG}{HTML}{F7E1E1}         %
\definecolor{SRBG}{HTML}{F3EEE8}          %
\definecolor{ColFact}{HTML}{2563EB}
\definecolor{ColReport}{HTML}{7C3AED}
\definecolor{ColCite}{HTML}{16A34A}
\definecolor{ColCross}{HTML}{DC2626}
\definecolor{QuestRow}{HTML}{EFF6FF}
\definecolor{Col2B}{HTML}{BFDBFE}
\definecolor{Col4B}{HTML}{93C5FD}
\definecolor{Col9B}{HTML}{60A5FA}
\definecolor{ColStage}{HTML}{1D4ED8}
\definecolor{promptonecolor}{HTML}{c0c0c0}
\definecolor{prompttwocolor}{HTML}{5d80fa}

\newcommand{\gsyn}{$G_{syn}$}
\newcommand{\gcondense}{$G_{cond}$}
\newcommand{\gtraj}{$G_{traj}$}

\newcommand{\legendsq}[1]{\textcolor{#1}{\rule[0.15ex]{1.1ex}{1.1ex}}\hspace{0.3em}}
\newcommand{\lbl}[4][]{\node[font=\tiny, color=black!65, #1] at (axis cs:#2,#3) {#4};}

\pgfplotsset{
bench bar/.style={
  width=4.4cm, height=3cm,
  ybar,
  bar width=11pt,
  bar shift=0pt,
  enlarge x limits=0.3,
  symbolic x coords={A,B,C,D},
  xtick=data, xticklabel=\empty,
  xtick style={draw=none},
  xlabel style={font=\scriptsize\bfseries, yshift=9pt},
  tick label style={font=\scriptsize, color=black!70},
  ymajorgrids=true, xmajorgrids=false,
  grid style={dashed, black!18, very thin},
  axis line style={black!18, thin},
  tick style={black!18},
  point meta=y,
  nodes near coords={\pgfmathprintnumber[fixed,precision=1]{\pgfplotspointmeta}},
  every node near coord/.append style={
    font=\fontsize{6}{6}\selectfont,
    color=black!75,
    anchor=south,
    inner sep=0pt,
    yshift=0.35pt,
  },
  clip=false,
},
bench line/.style={
  width=4.4cm, height=3.2cm,
  symbolic x coords={Vanilla,SFT,+MT,+RL},
  xtick={Vanilla,SFT,+MT,+RL},
  xmin=Vanilla, xmax=+RL,
  enlarge x limits=0.12,
  xticklabel style={font=\fontsize{6pt}{7pt}\selectfont, color=black!70, yshift=-1pt},
  xlabel style={font=\scriptsize\bfseries, yshift=2pt},
  tick label style={font=\scriptsize, color=black!70},
  grid=major, grid style={dashed, black!18, very thin},
  axis line style={black!18, thin},
  tick style={black!18},
}
}

\newtcolorbox{trainingexample}[2][]{%
  enhanced, breakable, colframe=black!12, colback=white, boxrule=2.5pt,
  arc=2pt, left=0pt, right=0pt, top=0pt, bottom=0pt,
  title={#2}, fonttitle=\bfseries, coltitle=black, #1}
\newtcblisting{promptbox}{%
  enhanced jigsaw,
  breakable,
  listing only,
  title=Quest Prompt,
  fonttitle=\bfseries\rmfamily\small,
  coltitle=white,
  colbacktitle=prompttwocolor,
  colback=prompttwocolor!10,
  colframe=prompttwocolor!10,
  boxrule=0pt,
  arc=2pt,
  outer arc=2pt,
  toptitle=2pt,
  bottomtitle=2pt,
  left=10pt,
  right=10pt,
  top=6pt,
  bottom=8pt,
  before skip=10pt,
  after skip=10pt,
  pad after break=4mm,
  listing options={%
    basicstyle=\small\ttfamily,
    breaklines=true,
    breakatwhitespace=true,
    columns=fullflexible,
    keepspaces=true,
    showstringspaces=false,
    extendedchars=true,
    inputencoding=utf8,
    aboveskip=0pt,
    belowskip=0pt,
    literate=%
      {–}{{-{}-}}1%
      {—}{{-{}-{}-}}1%
      {→}{{$\rightarrow$}}1%
      {←}{{$\leftarrow$}}1%
      {≤}{{$\leq$}}1%
      {≥}{{$\geq$}}1%
      {≠}{{$\neq$}}1%
      {…}{{...}}3%
      {‘}{{`}}1%
      {’}{{'}}1%
      {“}{{``}}1%
      {”}{{''}}1%
      {•}{{\textbullet}}1%
      {⸻}{{---}}3%
      {❌}{{$\times$}}1%
      {×}{{$\times$}}1,
  },
}

\newcolumntype{L}[1]{>{\RaggedRight\arraybackslash}p{#1}} %
\newcolumntype{Y}{>{\RaggedRight\arraybackslash}X}        %

\usepackage{etoolbox}
\appto\appendix{\section*{Appendix}}

\usepackage{blindtext}
\usepackage{float}
\usepackage{commath}
\usepackage{soul}
\usepackage{threeparttable}
\usepackage[colorinlistoftodos]{todonotes}
\usepackage{fontawesome5}

\newcommand{\ours}{\textsc{Quest}}
\newcommand{\capOne}{\textit{fact seeking}}
\newcommand{\capTwo}{\textit{citation grounding}}
\newcommand{\capThree}{\textit{report synthesis}}
\newcommand{\memoryState}{Context State}
\newcommand{\ourmodel}{\textsc{Quest-35B}}
\newcommand{\ourtb}{\textsc{Quest-30B}}

\title{\ours: Training Frontier Deep Research Agents \\ with Fully Synthetic Tasks}
\renewcommand{\titlelist}{%
  {\fontfamily{qpl}\selectfont\bfseries\fontsize{20}{26}\selectfont\color{osufg}%
   \ours: Training Frontier Deep Research Agents with Fully Synthetic Tasks}%
}
\shorttitle{\ours}
\shortauthors{Jian Xie et al.}

\author{Jian Xie}
\author{Tianhe Lin}
\author{Zilu Wang}
\author{Yuting Ning}
\author{Yuekun Yao}
\author{Tianci Xue}
\author{Zhehao Zhang}
\author{Zhongyang Li}
\author{Kai Zhang}
\author{Yufan Wu}
\author{Shijie Chen}
\author{Boyu Gou}
\author{Mingzhe Han}
\author{Yifei Wang}
\author{Vint Lee}
\author{Xinpeng Wei}
\author{Xiangjun Wang}
\author{Yu Su}
\author{Huan Sun}

\affiliation{The Ohio State University}
\affiliation{Amazon AGI SF Lab}

\newcommand{\core}[1]{{#1$^{*}$}}
\newcommand{\lead}[1]{{#1$^{*\dagger}$}}
\newcommand{\osu}[1]{#1$^{1}$}
\newcommand{\amazon}[1]{#1$^{2}$}
\newcommand{\authorunit}[1]{\mbox{#1}}
\renewcommand{\authorlist}{%
  {\sffamily\setstretch{1.55}%
  \authorunit{\osu{\lead{Jian Xie}}},
  \authorunit{\osu{\core{Tianhe Lin}}},
  \authorunit{\osu{\core{Zilu Wang}}},
  \authorunit{\osu{Yuting Ning}},
  \authorunit{\osu{Yuekun Yao}},
  \authorunit{\osu{Tianci Xue}},
  \authorunit{\osu{Zhehao Zhang}},
  \authorunit{\osu{Zhongyang Li}},
  \authorunit{\osu{Kai Zhang}},
  \authorunit{\osu{Yufan Wu}},
  \authorunit{\osu{Shijie Chen}},
  \authorunit{\osu{Boyu Gou}},
  \authorunit{\osu{Mingzhe Han}},
  \authorunit{\amazon{Yifei Wang}},
  \authorunit{\amazon{Vint Lee}},
  \authorunit{\amazon{Xinpeng Wei}},
  \authorunit{\amazon{Xiangjun Wang}},
   \authorunit{\osu{Yu Su}},
  \authorunit{\osu{Huan Sun}}
}}

\renewcommand{\affiliationlist}{%
  {\color{osufg}%
  {\normalsize\sffamily $^{1}$The Ohio State University \quad $^{2}$Amazon AGI SF Lab}\\[3pt]
  {\small
  {$^{*}$Core Contributors\enspace
  $^{\dagger}$Project Lead}}}%
}
\abstract{Deep research agents extend the role of search engines from retrieving keyword-matched pages to synthesizing knowledge, fundamentally changing how humans interact with information. 
However, frontier systems remain proprietary, while existing open agents often generalize poorly across different task types, leaving unclear how to train a broadly capable deep research agent.
\textbf{We release \ours{}, a family of open models (ranging from 2B to 35B) that serve as \textit{general-purpose} deep research agents designed to handle a wide range of long-horizon search tasks, with strong capabilities in \textit{fact seeking}, \textit{citation grounding}, and \textit{report synthesis}}.
To build \ours{}, we propose an effective training recipe combining mid-training, supervised fine-tuning, and reinforcement learning.
Central to this recipe is a curated data synthesis pipeline based on unified rubric trees, which applies to different task types and enables synthesizing training data with verifiable rewards without human annotation. 
In addition, \ours{} incorporates a built-in context management mechanism that enables effective long-horizon reasoning and knowledge synthesis.
\textbf{Using only 8K synthesized tasks, \ours{} approaches or even surpasses frontier closed-source agents across eight deep research benchmarks spanning diverse task types, and achieves the best overall performance among recent open-weight agents.}
We released everything: models, data, and training scripts. 
}
\newcommand{\resourcelink}[3]{%
  {\color{osufg}#1}\hspace{0.35em}{\sffamily\bfseries #2:}~%
  \href{#3}{\texttt{\textcolor{osured}{#3}}}\par
}
\newcommand{\hflogo}{%
  \smash{\raisebox{-0.2em}{\includegraphics[height=1.05em]{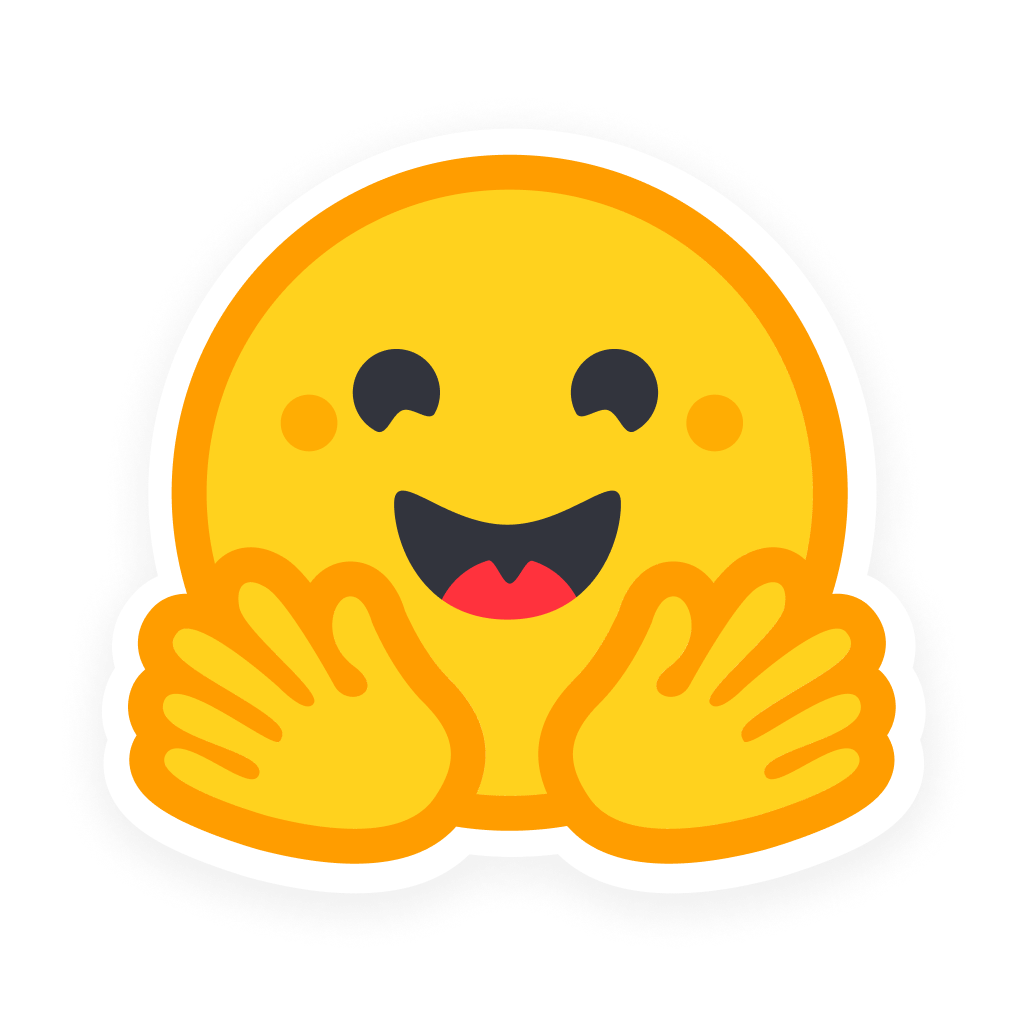}}}%
}
\renewcommand{\infolist}{%
  {\small\color{osufg}%
  {\color{osufg}\faEnvelope}\hspace{0.35em}{\sffamily\bfseries Correspondence:}~%
  \texttt{\textcolor{black}{\{xie.1741, su.809, sun.397\}@osu.edu}}\par
  \resourcelink{\faGlobe}{Project Page}{https://osu-nlp-group.github.io/QUEST}
  \resourcelink{\faGithub}{GitHub}{https://github.com/OSU-NLP-Group/QUEST}
  \resourcelink{\hflogo}{Demo}{https://huggingface.co/spaces/osunlp/QUEST}
  \resourcelink{\hflogo}{Model}{https://huggingface.co/osunlp/QUEST-35B-RL}}%
}

\begin{document}
\maketitle

\begin{figure*}[h]
\centering
\vspace{-2em}
    \includegraphics[width=\linewidth]{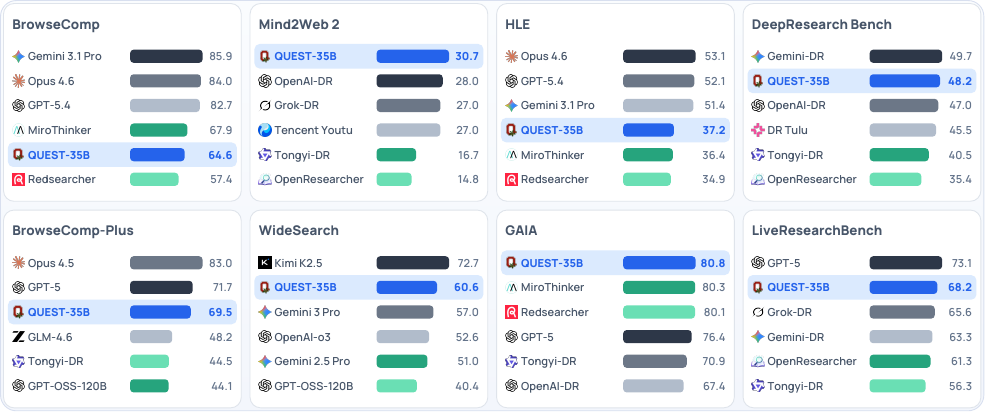}
    \vspace{-1em}
  \caption{
Comprehensive comparison across eight benchmarks that evaluate different capabilities: \capOne~(e.g., BrowseComp), \capTwo~(e.g., Mind2Web 2), and \capThree~(e.g., DeepResearch Bench). 
For BrowseComp and BrowseComp-Plus, following prior work~\citep{anthropic2025claude45opus, team2026kimi,chu2026redsearcher}, we adopt the discard-all strategy for context management. 
``-DR'' denotes Deep Research systems, such as OpenAI DeepResearch. 
Please refer to Appendix~\ref{app:evaluation-qwen35} for more details.}
    \label{fig:quest35b-results}
\vspace{-2em}
\end{figure*}

\section{Introduction}
\label{sec:intro}

\begin{figure*}[t]
\centering
    \includegraphics[width=\linewidth]{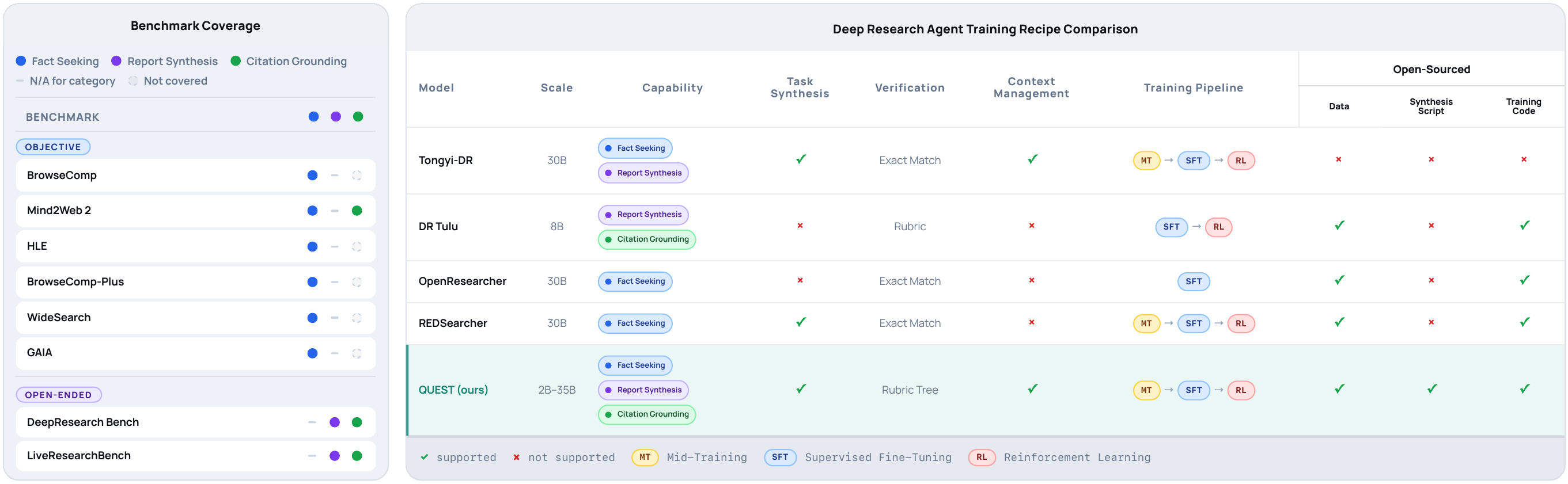}
    \caption{Comparison of training recipes for deep research agents. \ours{} covers
    \textit{fact seeking}, \textit{citation grounding}, and \textit{report synthesis} capabilities and provides a fully open recipe needed for reproducible training.}
    \label{fig:comparison}
\end{figure*}

Web search is moving from manual information gathering toward autonomous evidence-seeking and synthesis.
This shift builds on a progression of search interfaces: traditional search engines return ranked webpages for users to inspect, while retrieval-augmented generation (RAG) systems~\citep{lewis2020retrieval,gao2023retrieval} retrieve relevant documents and condition LLM responses on them.
Deep research agents~\citep{openai2025deepresearch, gemini_dr, team2025tongyi} push this paradigm further: given a complex information-seeking task, they can decompose it into intermediate goals, execute web queries, examine external sources, and synthesize citation-backed responses.
By shifting the burden of web-scale information seeking from humans to autonomous agents, these systems promise to make complex research workflows more efficient and scalable.
Yet building such agents is far from straightforward: they must learn not only to answer questions, but to search, verify, remember, and synthesize information over long horizons.
However, despite rapid progress, the training recipes enabling these capabilities remain poorly understood, as state-of-the-art systems often rely on proprietary models, datasets, and training pipelines with limited transparency.

Recent efforts have begun to address this gap by developing deep research agents using open-weight models, data, and training strategies~\citep{team2025tongyi, li2025websailor, li2026openresearcherfullyopenpipeline, shao2025drtulureinforcementlearning,du2026openseeker,du2026openseekerv2}. 
However, existing approaches typically focus on specific use scenarios and fail to capture the full range of capabilities required for general-purpose deep research agents. 
At the core, deep research tasks differ in how response quality can be assessed, spanning two regimes of correctness: \textbf{\emph{objective tasks}}, whose outputs can be grounded in externally verifiable evidence, and \textbf{\emph{open-ended tasks}}, whose outputs must be assessed through subjective judgment.

Based on this landscape, we identify three capabilities that together define a comprehensive profile for deep research agents.
Within objective tasks, one fundamental capability is \textbf{\capOne}, where the agent must locate a specific piece of information through multi-hop web search, as exemplified by BrowseComp~\citep{wei2025browsecomp}, which focuses on obscure facts and thus makes fact seeking especially challenging.
Open-ended tasks require a further capability, \textbf{\capThree}, as exemplified by DeepResearch Bench~\citep{du2025deepresearch}.
Given an open-ended task, the agent must synthesize information from diverse sources into a coherent, well-structured, and reader-friendly report.
Here, evaluation relies primarily on rubric-based judgment of coverage, organization, and clarity, rather than on a single externally verifiable answer.
Another shared capability across both objective tasks and open-ended tasks is \textbf{\capTwo}: the agent must support its claims with reliable, up-to-date, and verifiable citations. 
Mind2Web 2~\citep{gou2025mind2web2} exemplifies this setting, as its evaluation relies on explicit constraints and citation-backed verification. 
Together, these three capabilities define a broad capability profile for deep research agents: retrieving precise information, synthesizing knowledge into coherent, well-structured reports, and providing verifiable citations to support the claims in their responses.
Despite the complementary nature of these three capabilities, existing benchmarks and agent systems evaluate or support them only in isolation, as illustrated in Figure~\ref{fig:comparison}, and no prior agent system addresses them jointly within a unified framework.

In this paper, we introduce \ours{}, a family of open state-of-the-art models ranging from 2B to 35B parameters, designed as general-purpose deep research agents.
The advancements of \ours\ are attributed to three technical aspects: a novel \textbf{data synthesis pipeline} for constructing training tasks paired with tree-structured rubrics, a \textbf{context management strategy} that maintains compact summaries of information throughout long-horizon trajectories, and an effective \textbf{training pipeline} that optimizes the model using rubric-based signals.
To comprehensively evaluate \ours{} across different deep research capabilities, we adopt eight diverse benchmarks covering \capOne{}, \capThree{}, and \capTwo{}.
Across these benchmarks, \ourmodel{} shows consistently strong performance (See Figure~\ref{fig:quest35b-results}), e.g., reaching 64.6\% on BrowseComp, 30.7\% on Mind2Web 2, and 48.2\% on DeepResearch Bench, surpassing strong proprietary systems such as OpenAI DeepResearch.

Our contributions are as follows:
\textit{1)} A scalable data synthesis pipeline for deep research agent training, which produces \ours{}-8K, a high-quality synthetic dataset where each instance pairs a complex query with a task-specific, verifiable rubric tree.
\textit{2)} A context management strategy for long-horizon deep research tasks, maintaining compact summaries of accumulated information across extended search trajectories.
\textit{3)} An effective training pipeline that combines mid-training, supervised fine-tuning, and reinforcement learning, using task-specific rubrics as optimization signals.
\textit{4)}  Controlled studies that analyze the contribution of each training stage and the scaling behavior across model sizes (2B, 4B, 9B, up to 35B parameters).
We find that even small models (e.g., \ours{}-2B-SFT and \ours{}-4B-SFT) achieve impressive performance, showing the potential of small-scale research agents locally deployable for privacy-sensitive settings, such as biomedical research and legal analysis.

\section{Data Synthesis}
Deep research training data should contain challenging queries that require long-horizon search, webpage visiting, and reasoning. 
Existing deep research agents~\citep{li2025websailor, chu2026redsearcher} largely synthesize such data as complex questions with a single verifiable answer.
This format is well-suited to BrowseComp-like tasks, e.g., ``\textit{Which architect designed the house where the author of The Catcher in the Rye lived for decades?}'' where the answer is a hard-to-find entity.

However, a general-purpose deep research agent requires training data beyond such single-answer supervision.
First, training on such data does not generalize to the broad range of tasks that deep research agents may encounter.
Single-answer supervision mainly targets fact seeking tasks with short final answers, whereas many deep research tasks require aggregating and reconciling information across multiple sources into coherent, source-supported outputs.
Second, single-answer queries typically induce a binary correctness reward for reinforcement learning, which limits effective credit assignment.
In contrast, deep research tasks often involve multiple evaluation criteria, such as source attribution and insightfulness.
Exploiting such fine-grained signals as rewards can guide more effective optimization.
We next describe how our synthesis pipeline addresses them.

\forestset{
  rubric/.style={
    draw=gray!55, rounded corners=4pt,
    font=\scriptsize,
    inner xsep=5pt, inner ysep=4pt,
    edge={draw=gray!45, line width=0.45pt, -latex},
    l sep=12mm, s sep=5mm,
  },
  rubric horiz/.style={
    grow'=east,
    parent anchor=east, child anchor=west,
    draw=gray!55, rounded corners=4pt,
    font=\scriptsize,
    inner xsep=5pt, inner ysep=4pt,
    edge={draw=gray!45, line width=0.45pt, -latex},
    l sep=8mm, s sep=2mm,
    anchor=west,
  },
  branch/.style={fill=ColFact!10, draw=ColFact!55, line width=0.5pt},
  leaf/.style={fill=ColCross!7, draw=ColCross!45, line width=0.4pt},
  root/.style={fill=gray!12, draw=gray!60, line width=0.55pt},
}

\begin{figure}[t]
\centering
\includegraphics[width=\linewidth]{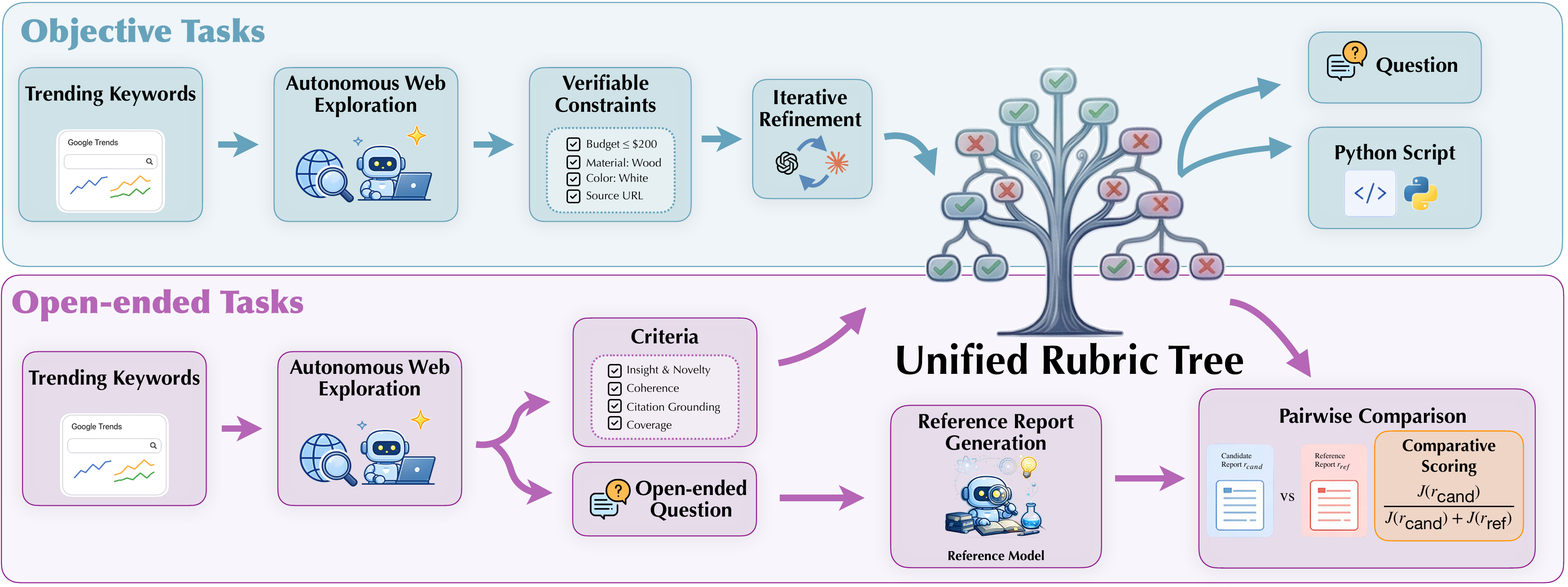}

\vspace{0.6em}

\begin{subfigure}[t]{0.46\linewidth}
\centering

\begin{tcolorbox}[
  colback=blue!3,
  colframe=blue!55!black,
  boxrule=0.5pt,
  arc=2pt,
  left=4pt,right=4pt,top=3pt,bottom=3pt,
  width=\linewidth
]
\scriptsize
\textbf{Query:} Identify the two 2024 U.S.\ listeria outbreaks that caused deaths, and determine which was deadlier.
\end{tcolorbox}

\vspace{2.5em}

\begin{minipage}[b][3.2cm][c]{\linewidth}
\centering
\resizebox{\linewidth}{!}{%
\begin{forest}
  forked edges,
  for tree={rubric horiz}
  [{\parbox[c]{1.7cm}{\centering\textbf{Root}\\\scriptsize Listeria outbreaks}}, root
    [{\parbox[c]{2.2cm}{\centering\textbf{Outbreak 1}\\\scriptsize Boar's Head}}, branch
      [{\parbox[c]{2.4cm}{Source: Boar's Head deli meats}}, leaf]
      [{\parbox[c]{2.4cm}{Death toll: 10}}, leaf]
    ]
    [{\parbox[c]{2.2cm}{\centering\textbf{Outbreak 2}\\\scriptsize Rizo-L\'opez}}, branch
      [{\parbox[c]{2.4cm}{Source: queso fresco / cotija}}, leaf]
      [{\parbox[c]{2.4cm}{Death toll: 2}}, leaf]
    ]
    [{\parbox[c]{2.2cm}{\centering\textbf{Comparison}}}, branch
      [{\parbox[c]{2.4cm}{Concludes Boar's Head was deadlier}}, leaf]
      [{\parbox[c]{2.4cm}{Cites 10 vs.\ 2 deaths as evidence}}, leaf]
    ]
  ]
\end{forest}
}
\end{minipage}
\vspace{0.65em}
\caption{Rubric tree of an objective task}
\label{fig:rubric-objective}
\end{subfigure}
\hfill
\begin{subfigure}[t]{0.53\linewidth}
\centering

\begin{tcolorbox}[
  colback=blue!3,
  colframe=blue!55!black,
  boxrule=0.5pt,
  arc=2pt,
  left=4pt,right=4pt,top=3pt,bottom=3pt,
  width=\linewidth
]
\scriptsize
\textbf{Query:} Assess Apple's entry into foldable smartphones, focusing on technical comparison, adoption barriers, pricing, and presentation.
\end{tcolorbox}

\vspace{2em}

\begin{minipage}[b][3.2cm][c]{\linewidth}
\centering
\resizebox{\linewidth}{!}{%
\begin{forest}
  forked edges,
  for tree={rubric horiz}
  [{\parbox[c]{1.7cm}{\centering\textbf{Root}\\\scriptsize Foldable assessment}}, root
    [{\parbox[c]{2.6cm}{\centering\textbf{Comprehensiveness}}}, branch
      [{\parbox[c]{4.2cm}{Compare vs.\ Samsung Z Fold}}, leaf]
      [{\parbox[c]{4.2cm}{Compare vs.\ Huawei Mate X}}, leaf]
      [{\parbox[c]{4.2cm}{Cover hinge / display supply chain}}, leaf]
    ]
    [{\parbox[c]{2.6cm}{\centering\textbf{Insight}}}, branch
      [{\parbox[c]{4.2cm}{Analyze adoption barriers}}, leaf]
      [{\parbox[c]{4.2cm}{Discuss market-timing rationale}}, leaf]
    ]
    [{\parbox[c]{2.6cm}{\centering\textbf{Instruction}\\\textbf{Following}}}, branch
      [{\parbox[c]{4.2cm}{Address pricing strategy}}, leaf]
      [{\parbox[c]{4.2cm}{Address technical comparison}}, leaf]
    ]
    [{\parbox[c]{2.6cm}{\centering\textbf{Readability}}}, branch
      [{\parbox[c]{4.2cm}{Clear hierarchical structure}}, leaf]
      [{\parbox[c]{4.2cm}{Concise, jargon-free prose}}, leaf]
    ]
  ]
\end{forest}%
}
\end{minipage}
\vspace{1.1em}
\caption{Rubric tree of an open-ended task}
\label{fig:rubric-openended}
\end{subfigure}
\caption{
Data synthesis pipeline of \ours{}.
The top part illustrates the data generation pipeline.
The bottom shows rubric-tree examples for an objective task and an open-ended task, together with corresponding queries.
\textbf{Gray} nodes mark the root, \textbf{blue} nodes are intermediate branches that group sub-criteria, and \textbf{red} nodes are leaf-level constraints that can be directly verified against the response.
}
\label{fig:task}
\end{figure}

\subsection{Rubric Tree} \label{sec:data:rubric}
At the core of our synthesis pipeline is a \textit{rubric tree} introduced in Mind2Web 2~\citep{gou2025mind2web2}: a hierarchical decomposition of the constraints that a valid answer should satisfy.
Unlike Mind2Web 2, which relies on human-written tasks and human-refined evaluation scripts, \ours{} constructs rubric trees fully synthetically through automatic generation followed by strict refinement and verification, making large-scale data synthesis possible.
As shown at the bottom of Figure~\ref{fig:task}, the root node represents the overall score, which is aggregated from its child nodes, each corresponding to a task-specific criterion.
Each leaf node represents a directly verifiable criterion, such as factual correctness or source attribution, and receives a binary score through automatic verification.
Each internal node represents a higher-level constraint that is recursively decomposed into finer-grained child nodes, and receives a score aggregated from its children.
We describe how rubric trees are automatically constructed in Section~\ref{sec:data:obj} and show complete examples in Appendix~\ref{app:obj-example}.

This design naturally addresses the limitations of answer-centric QA supervision.
By representing task-specific constraints rather than a unique ground-truth answer, rubric trees can accommodate tasks with unique answers, tasks with multiple valid solutions, and open-ended tasks requiring multi-aspect evaluation.
This allows us to generate diverse training data under a unified framework.
Moreover, the partial score of the root node in the rubric tree provides a fine-grained training signal beyond binary correctness, reflecting the degree to which a prediction satisfies the underlying constraints.

\subsection{Data Generation Pipeline} \label{sec:data:obj}
Our rubric-tree-centric pipeline generates both objective and open-ended training data.
Each instance consists of a complex query (i.e., a task description), a query-specific rubric tree, and an evaluation protocol. 
Given an agent's response to the query, the evaluation protocol maps the response to a score in \([0,1]\) according to the rubric tree.
Figure~\ref{fig:task} presents an overview of our pipeline.
We denote by $G_{\mathrm{syn}}$ the capable LLM used for both task generation and rubric-tree generation throughout the pipeline (i.e., Claude Sonnet 4.5 unless otherwise specified).

\paragraph{Objective Tasks.}
We first sample trending keywords from a pool crawled from Google Trends as topical seeds for downstream synthesis, ensuring that the resulting tasks remain both topically relevant and temporally diverse, while also better reflecting real-world user needs.
For each set of trending keywords, we prompt \gsyn\ to autonomously browse the web, collect relevant information, derive a set of verifiable constraints from the retrieved content, and organize them into a rubric tree.
All rubric trees undergo iterative refinement and verification to ensure a correct tree structure and valid node definitions.
Judged by Claude Sonnet 4.5, tasks whose rubric trees still cannot be resolved into a consistent, reliably evaluable structure are discarded.

Given a rubric tree, we first prompt \gsyn\ to translate it into a natural-language question, and then use GPT-5 to generate an executable Python evaluation script as the corresponding evaluation protocol. 
The script programmatically verifies each rubric node against the agent’s response and computes the final score, enabling fully automated evaluation at scale.
We provide all the prompts used for synthesis and the details of human examination in Appendix~\ref{app:task-synthesis-prompts} and Appendix~\ref{app:python-script-check}.

\paragraph{Open-ended Tasks.}
Open-ended tasks basically follow the same keywords sampling and web exploration process, but differ in how rubrics are constructed. 
Instead of using fully task-specific rubric trees, we fix 
the children of the root node
across examples with four shared criteria following DeepResearch Bench~\citep{du2025deepresearch}: instruction following, comprehensiveness, readability, and insight. 
The child nodes under each criterion are adaptive and contain task-specific nodes generated by \gsyn\ conditioned on the question.
For score aggregation, we assign a weight to each node using \gsyn. 
To improve stability, we generate the weights three times and average them across runs.

The evaluation protocol of an open-ended task consists of three components: a reference report, a rubric-based judge, and a pairwise normalization rule.
Given a task, we first prompt \gsyn\ to produce a reference report \(r_{\mathrm{ref}}\).
Then, for each candidate report \(r_{\mathrm{cand}}\), a judge model receives both the candidate and reference reports and scores them under each task-specific rubric node.
A judge model then receives both the candidate report generated by the tested model and the reference report simultaneously---seeing both at once allows the judge to make finer-grained quality distinctions than independent scoring permits.
For each task-specific rubric node, the judge assigns separate continuous scores from 0 to 10 to the candidate report and the reference report, where higher scores indicate better satisfaction of the corresponding criterion.
The node-level scores are aggregated through the rubric tree, yielding weighted scores \(J(r_{\mathrm{cand}})\) and \(J(r_{\mathrm{ref}})\), where \(J(\cdot) \in [0,1]\).
The final score is defined as \(\mathrm{Score} = J(r_{\mathrm{cand}}) / (J(r_{\mathrm{cand}}) + J(r_{\mathrm{ref}}))\), which measures the candidate report's quality relative to the reference report.
A score above 0.5 indicates that the candidate has surpassed the quality of the reference report.

\begin{figure*}[t]
\centering
    \includegraphics[width=\linewidth]{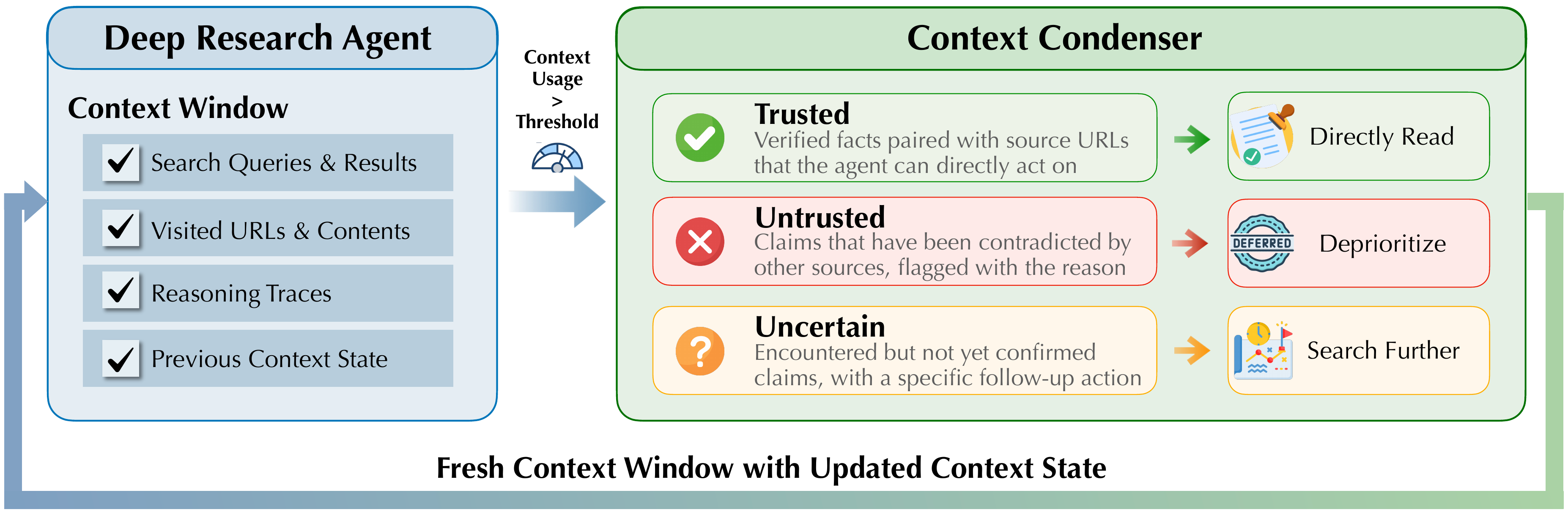}
    \caption{Context management in \ours{}. When the context exceeds a threshold, a Context Condenser compresses the history into a structured \memoryState{} with trusted, untrusted, and uncertain entries, enabling \ours{} to resume in a fresh context window.}
    \label{fig:memory}
\end{figure*}

\section{Context Management} 
\label{sec:memory}
A deep research agent searches, reads, and revises through multi-turn tool use before producing a final answer.
As the context window fills with raw search results, visited pages, and intermediate reasoning traces, the agent's ability to attend to what matters most begins to degrade.
Existing open-weight agents sidestep this problem by either capping the number of turns or relying on a sufficiently large context window to hold everything at once.
By contrast, \ours{} adopts a structured context management module that compresses all the history context into a compact summary, enabling the agent to reason over arbitrarily long horizons without context overflow.

Specifically, we use a structured JSON object, called \memoryState, to store the agent's history context.
The \memoryState\ organizes accumulated knowledge into three buckets. 
\textit{Trusted} entries contain facts that the agent has previously checked against retrieved source URLs based on its reasoning trace, allowing them to be directly reused without additional verification.
\textit{Untrusted} entries are claims contradicted by other sources, flagged with the reason for distrust.
\textit{Uncertain} entries are partially supported claims that require follow-up, each annotated with a URL to visit or a query to re-run.
This structured state encodes the agent's epistemic state,
allowing the resumed agent to distinguish between what it knows, doubts, and still needs to verify.

During an agent trajectory, as shown in Figure~\ref{fig:memory}, context management is triggered once the context window usage exceeds a threshold, using a \textit{Context Condenser} model \gcondense.
We instantiate \gcondense\ with GPT-5-mini; see Appendix~\ref{app:inference-prompts} for the prompt.
The condenser takes as input the full raw history, which includes search queries and results, visited URLs with extracted content, reasoning traces, and any previously summarized memory, and then produces an updated \memoryState.
After condensation, \ours{} resumes in a fresh context window initialized with the updated \memoryState.
Uncertain entries guide subsequent search and verification actions, while untrusted entries are deprioritized and not actively reused unless further verification is required.
By reusing previously verified claims without redundant tool calls, the structured \memoryState{} enables coherent, citation-grounded knowledge synthesis across long-horizon research tasks.

\section{Training}
\ours{} training pipeline includes three stages: mid-training (MT),
supervised fine-tuning (SFT), and reinforcement learning (RL). 
Each stage plays a complementary role in shaping agent behaviors: MT adapts the base model to long-horizon search interaction, SFT teaches the agent to imitate high-quality tool-use trajectories, and RL optimizes the policy using rubric-based and fact-checking rewards.

\subsection{Trajectory Collection} \label{sec:training:traj}
The SFT training requires full agent trajectories for queries generated from our data synthesis pipeline.
We collected such trajectories by running a teacher model \gtraj\ on the synthesized task set, where the model iteratively interacts with tools in our tool set and conditions on the returned observations.
We use Tongyi DeepResearch~\citep{team2025tongyi} as \gtraj, and use GPT-5.2 to polish the final report for open-ended tasks.

Given a training query, we prompt \gtraj\ to solve it and evaluate the output with the query-specific evaluation protocol.
If the score meets or exceeds a threshold \(\epsilon\), we retain the generated trajectory as an SFT target for this query.
For objective tasks, where the initial trajectory does not meet this threshold, we apply a reflection-based retry strategy~\citep{zhang2025ee}: the fine-grained evaluation result from the protocol is injected into the prompt as a hint, and \gtraj\ is prompted to attempt the query again.
The retried trajectory is retained if its evaluation score meets or exceeds \(\epsilon\).
We set $\epsilon=1$ and $0.475$ for objective tasks and open-ended tasks, respectively.

\textbf{Inline Citation Format.}
To improve the readability and verifiability of generated reports, we standardize all outputs into an inline citation format, where each factual claim is directly annotated with the URL of its supporting source.
Specifically, for each trajectory, we employ GPT-5-mini to retrospectively insert inline citations into the original response, drawing on both the reasoning trace and the \memoryState{}---which records the full history of visited web content and extracted information.
This also establishes the citation grounding output structure that the fact-checking reward in the RL stage will later reinforce.

\vspace{-0.5em}
\subsection{Mid-Training}
The MT stage consists of two auxiliary tasks that equip the base model with long-context understanding and awareness of the \memoryState{} structure, which the agent relies on in later stages.

\textbf{\textit{1)} Context Summarization.}
Given a long history context, the model is required to produce a structured \memoryState{} JSON of the form described in Section~\ref{sec:memory}.
The summaries used as supervision targets are produced by GPT-5-mini during data synthesis and reused here directly, avoiding additional annotation cost while ensuring that the \memoryState{} format is consistent with the agent process during inference.
\textbf{\textit{2)} Relevant Information Extraction.}
Given a raw HTML page and an extraction goal, the model is required to produce a goal-relevant summary that filters out navigation elements, advertisements, and off-topic content.
This trains the model to attend selectively within long, noisy contexts, a capability that is prerequisite to effective web-based research but largely absent in models trained on clean document corpora.

Both tasks require no additional data collection.
The relevant information extraction targets are derived directly from the 
output of the visit tool from trajectories in Section~\ref{sec:training:traj},
and the context summarization targets are produced by \gcondense\ throughout the pipeline, making MT a natural byproduct of the overall framework rather than an annotation effort.
For more details, please refer to Appendix~\ref{app:data-mt}.

\subsection{Supervised Fine-Tuning}
Following MT, the SFT stage trains the model on trajectories from Section~\ref{sec:training:traj} using the next-token prediction objective.
Given a full training trajectory, we first transform it into multiple session-level training instances for efficient optimization. 
We define a session as a contiguous segment of interaction between two consecutive context condensation events. 
Concretely, each session contains all tokens from the current working context after the previous context condenser is applied, up to (but not including) the next condensation step. 
Thus, each completed trajectory is divided into multiple sessions, depending on how many times the context condenser is triggered.
This design aligns the training unit with the agent's effective working context during inference.
We provide more discussion about the advantages of session-level training in Appendix~\ref{app:session-level-training}.

\subsection{Reinforcement Learning}
In the RL stage, we use GRPO-style outcome-based reinforcement tuning~\citep{guo2025deepseekr1}.
The reward function combines the rubric-tree reward with the fact-checking reward defined below.
We follow the same session-level training strategy as in the SFT stage.

\paragraph{Rubric-tree Rewards.}
The rubric-tree reward $s_{\mathrm{rubric}}$ of an agent's response 
is derived from the evaluation score calculated by the query-specific evaluation protocol. 
For objective tasks, we directly use this score as $s_{\mathrm{rubric}}$.
For open-ended tasks, the protocol produces a raw pairwise score comparing the agent response with a reference response, which we calibrate into ordered discrete reward levels by mapping higher pairwise scores to larger rewards.
In our implementation, pairwise scores at least 0.5 are assigned $s_{\mathrm{rubric}}=1.0$; scores in [0.475, 0.5), [0.45, 0.475), and [0.425, 0.45) are assigned $0.75$, $0.5$, and $0.25$, respectively; and scores below 0.425 are assigned $0$.
This preserves the relative ordering of responses while making the reward signal more separable and less sensitive to small score fluctuations.

\paragraph{Fact-checking Rewards.}
For both objective and open-ended tasks, we add a fact-checking reward $s_\mathrm{fact}$ to encourage the agent to ground its responses in verifiable sources through its inline citations.
For each response, we first extract cited fact--URL pairs 
from the answer, de-duplicate them, retrieve the referenced webpages, and use an LLM-based evaluator (i.e.\ GPT-5-mini) to label each citation as \textit{supported}, \textit{unsupported}, or \textit{unknown}.
We then calculate $s_{\mathrm{fact}}$ as the fraction of supported citations among citations with determinate labels, i.e., those labeled as either {supported} or {unsupported}.
The final reward $R$ is calculated by:
\begin{equation}
\small
    R = 0.75 \cdot s_{\mathrm{rubric}} + 0.25 \cdot s_{\mathrm{fact}} \cdot s_{\mathrm{rubric}},
\end{equation}
where $s_{\mathrm{rubric}}, s_{\mathrm{fact}} \in [0,1]$.
The fact-checking term is scaled by the rubric-tree reward, so that well-supported citations cannot compensate for poor task completion, and the fact-checking contribution vanishes when the task is entirely failed.
Responses without inline citations receive $s_{\mathrm{fact}}=0$.

\paragraph{Reward and Advantage Propagation.}
For each input prompt, we sample a group of rollouts.
For each rollout $i$, we compute a scalar reward $R_i$ based on its full response and assign this reward to all sessions derived from the same rollout.
The advantage for each rollout is computed by normalizing its reward within the group:
\begin{equation}
\small
    A_i = \frac{R_i - \mu_g}{\sigma_g + \epsilon},
\end{equation}
where $\mu_g$ and $\sigma_g$ are the mean and standard deviation computed over unique rollouts in group $g$.
When the context condenser produces multiple sessions per rollout, these are deduplicated during normalization so that all sessions derived from the same rollout share a single advantage value.
We then optimize the policy using the standard GRPO objective~\citep{shao2024deepseekmath}, excluding the KL penalty.
\section{Infrastructure}
\label{app:infra}
During training and inference, \ours{} is equipped with a practical tool set to interact with external environments. Specifically, the tool set includes:
\textit{1)} \texttt{Google Search} for retrieving relevant online results for a given query;
\textit{2)} \texttt{Visit}, which reads webpages and summarizes information conditioned on the goal of extracting key information from the HTML;
\textit{3)} \texttt{Python Interpreter} for computation;
\textit{4)} \texttt{Google Scholar} for accessing academic publications.
For the summarization model used in the \texttt{Visit} tool, we adopt GPT-5-mini.

While these tools are essential for the agent's capabilities, training a deep research agent at scale introduces infrastructure challenges that go beyond standard LLM post-training. 
The agent must interact with live web content throughout data synthesis, SFT trajectory collection, and RL rollout---each of which involves hundreds of search queries and URL visits executed in parallel. 
Without careful infrastructure design, these external calls become prohibitively expensive and slow.
\ours{} addresses this through a dual-cache system that intercepts all search and visit operations before they reach the live API. 
Repeated queries and URL visits during data construction, SFT, and RL rollout are served from cache rather than billed as API calls. 
\paragraph{Search and Scholar Cache.}
We build the search and scholar cache online during data synthesis and training: whenever a live search API call is made, the query and returned results are written into a persistent cache for future reuse.
Every new search query first undergoes exact-match lookup against this cache.
On a cache hit, results are returned immediately without an API call. 
On a miss, the query is passed to the FAISS retrieval system~\citep{johnson2019billion} for semantic similarity search: if a sufficiently similar query with cosine similarity above threshold already exists in the cache, its results are returned as a proxy. 
Only when both exact and semantic lookup fail does the system fall through to the live search API, at which point the result is written back to the cache for future use. 
We use Serper as the live API.

\paragraph{Visit Cache.}
Visit tool follows a simpler exact-match policy. 
Each URL is checked against a persistent cache keyed by the full URL string. 
On a hit, the stored page content is returned directly. 
On a miss, the live page accessed is fetched and cached. 
This eliminates redundant fetches of the same page across different training runs.
We use Jina as the live API.

\paragraph{Fully Asynchronous Rollout, Evaluation, and Training for RL.}
Built on top of VERL's fully asynchronous policy optimization framework~\citep{sheng2024hybridflow}, which decouples rollout and training, we further extend the pipeline by making evaluation fully asynchronous.
Specifically, beyond overlapping the Rollouter and the Trainer, we offload time-consuming task-specific evaluation to a pool of asynchronous Ray actors dedicated to executing the evaluation function.
This extension is particularly important for rubric-tree-based evaluation. 
Unlike traditional rule-based rewards that can often be computed via simple heuristic checks, rubric-tree evaluators require structured assessment over multiple criteria nodes, leading to an average evaluation time of around 4 minutes and up to 30 minutes in long-tail cases in our tests.
To mitigate this, once a trajectory is generated, it is immediately dispatched to the evaluation pool. 
Instead of stalling the pipeline until an entire generation batch is scored, the training process continuously pulls evaluated samples from a shared queue. 
A model update is triggered asynchronously as soon as enough scored samples are accumulated to form a training batch. 
Consequently, slower evaluations from earlier rollouts are naturally absorbed into subsequent training steps alongside faster recent samples.
To support stable training, we further introduce evaluator timeout control and interruption-aware reward handling, allowing in-flight evaluations to be resumed or safely reissued during parameter synchronization.
Overall, this design reduces pipeline bubbles caused by both long-tail rollouts and long-tail evaluators, improving the efficiency of RL training.
\newpage\section{Experiments}
\label{sec:exp}

\subsection{Setup}
\paragraph{Training.}
Given that most existing open deep research agents chose a base model with around 30B parameters, we take Qwen3.5-35B-A3B~\citep{qwen3.5} as our default base model, and apply our full training recipe to obtain \ourmodel{}.
We use 8K instances generated through our data synthesis pipeline for SFT and RL training in total. 
Table~\ref{tab:data_dist} presents the data distributions across different training stages.
Please refer to Appendix~\ref{app:data} and~\ref{app:training-details} for more details.

\begin{figure*}[t]
\centering
\begin{minipage}[t]{0.45\linewidth}
\centering
\captionof{table}{Data distribution across different training stages. C.S. denotes the context summarization task, and R.I.E. denotes the relevant information extraction task. In MT, task, trajectory, and session are equivalent due to the single-turn setting.
}
\label{tab:data_dist}
\resizebox{\linewidth}{!}{%
\begin{tabular}{llrrr}
\toprule
\multirow{2}{*}{\textbf{Stage}} & \multirow{2}{*}{\textbf{Type}} 
  & \multicolumn{3}{c}{\textbf{Data Category}} \\
\cmidrule(lr){3-5}
& & \textbf{\#Task} & \textbf{\#Traj.} & \textbf{\#Session} \\
\midrule
\multirow{2}{*}{MT}
  & C.S.       & \multicolumn{3}{c}{309,346}     \\
  & R.I.E.   & \multicolumn{3}{c}{1,052,663}   \\
\midrule
\multirow{2}{*}{SFT}
  & Objective  & 5,070  & 19,435 & 39,861 \\
  & Open-ended & 1,958  &  4,485 & 11,903 \\
\midrule
\multirow{2}{*}{RL}
  & Objective  & 864  & --    & --   \\
  & Open-ended & 269    & --    & --    \\
\bottomrule
\end{tabular}
}
\end{minipage}
\hspace{0.03\linewidth}
\begin{minipage}[t]{0.5\linewidth}
\centering
\captionof{table}{Benchmark sizes (full set and subset) and evaluation metrics for the benchmarks used in \ours{}. Acc.\ denotes accuracy, SR denotes success rate, and Overall denotes the average across all evaluation metrics. For DRB, we report the RACE score.}
\label{tab:subset}
\resizebox{\linewidth}{!}
{%

\begin{tabular}{l c c}

\toprule

\textbf{Benchmark} & \textbf{Size (Full / Subset)} & \textbf{Criterion} \\

\midrule

BrowseComp         & 1,266 / 130 & Acc. \\
Mind2Web 2         & 120 / --    & SR \\
HLE-Text           & 2,158 / 130 & Acc. \\
DeepResearch Bench & 100 / --    & Overall \\
BrowseComp-Plus    & 830 / 130   & Acc. \\
WideSearch         & 200 / 100   & Acc. \\
GAIA-Text          & 103 / --    & Acc. \\
LiveResearchBench  & 100 / --    & Overall \\
\bottomrule

\end{tabular}
}
\end{minipage}
\end{figure*}

\paragraph{Benchmark Evaluation.}
We evaluate \ours{} on eight deep research benchmarks: six objective benchmarks, including BrowseComp (BC;~\citep{wei2025browsecomp}), Mind2Web 2 (M2W2;~\citep{gou2025mind2web2}), Humanity's Last Exam (HLE;~\citep{phan2025humanity}),  BrowseComp-Plus (BC-Plus;~\citep{chen2025BrowseCompPlus}), WideSearch~\citep{wong2025widesearchbenchmarkingagenticbroad}, and GAIA~\citep{mialon2023gaia}, and two open-ended benchmarks, DeepResearch Bench (DRB;~\citep{du2025deepresearch}) and LiveResearchBench (LRB;~\citep{wang2026liveresearchbenchlivebenchmarkusercentric}).
For the results in Figure~\ref{fig:quest35b-results}, we use the official full test sets, except for HLE and GAIA, where we evaluate on the text-only versions.
Unless otherwise specified, ablation studies and analyses of BC, HLE, WideSearch, and BC-Plus are conducted on sampled subsets due to the high cost of full set evaluation.
For details on the subset splits of the benchmarks, please refer to Table~\ref{tab:subset} and Appendix~\ref{app:subset}.

\paragraph{Baselines.}
We compare \ourmodel\ against prior deep research agents in two categories:
(1) \textit{Frontier proprietary agents},
including OpenAI DeepResearch (OpenAI-DR)~\citep{openai2025deepresearch}, GPT-5~\citep{singh2026openaigpt5card}, Claude Opus 4.5, and Gemini 3 Pro; and (2) \textit{open-weight agents at a similar scale}, including Tongyi DeepResearch (Qwen3-30B-A3B)~\citep{team2025tongyi}, 
OpenResearcher (Nemotron-3-Nano-30B-A3B)~\citep{li2026openresearcherfullyopenpipeline}, and DR Tulu (Qwen3-8B)~\citep{shao2025drtulureinforcementlearning}, where parentheses indicate the base models.
Since work for these models reports results on only a subset of our chosen benchmarks, we run Tongyi-DR and OpenResearcher ourselves to fill in missing numbers for a complete comparison.
For DR Tulu, we evaluate it on Mind2Web 2 ourselves and find that it is good at open-ended tasks that focus on report synthesis, but not at objective tasks, so we do not further evaluate it on other benchmarks.
Although RedSearcher~\citep{chu2026redsearcher} is claimed to be open-weight, its model weights were not publicly available at the time of our experiments; we therefore do not report its results.

\begin{table}[t]
\caption{Comparison of proprietary and open deep research agents across eight benchmarks. Results marked with $^{\dagger}$ are obtained from our implementation using the official evaluation scripts on the full set. $^{*}$ denotes results obtained using the discard-all strategy~\citep{anthropic2025claude45opus,team2026kimi, chu2026redsearcher} on the full set for fair comparison with proprietary agents. The highest score is marked in bold, and the second is underlined.}
\vspace{-1em}
\label{tab:main_results}
\centering
\resizebox{\linewidth}{!}{%
\begin{tabular}{lcccccc|cc}
\toprule
\textbf{Model} 
& \multicolumn{6}{c|}{\textbf{Objective Tasks}}
& \multicolumn{2}{c}{\textbf{Open-Ended Tasks}} \\
\cmidrule(lr){2-7} \cmidrule(lr){8-9}
& \textbf{BC} 
& \textbf{BC-Plus} 
& \textbf{M2W2} 
& \textbf{WideSearch} 
& \textbf{HLE} 
& \textbf{GAIA} 
& \textbf{DRB} 
& \textbf{LRB} \\
\midrule

\multicolumn{9}{c}{\textbf{Frontier Proprietary Agents}} \\
\midrule
GPT-5 
& 59.9 & \underline{71.7} & - & 54.0 & 35.2 & \underline{76.4} & - & \underline{73.1} \\
Claude Opus 4.5
& \textbf{67.8} & \textbf{83.0} & - & \textbf{76.2} & \underline{43.2} & - & \textbf{50.6} & - \\
Gemini 3 Pro 
& 59.2 & - & - & 57.0 & \textbf{45.8} & - & \underline{49.6} & - \\
OpenAI-DR 
& 51.5 & - & 28.0 & - & 26.6 & 67.4 & 47.0 & - \\

\midrule
\multicolumn{9}{c}{\textbf{Open-Weight Agents}} \\
\midrule
\rowcolor{orange!15}
Tongyi-DR 
& 43.4 & 44.5 & 16.7$^{\dagger}$ & 37.3$^{\dagger}$ & 32.9 & 70.9 & 40.5 & 56.3$^{\dagger}$ \\
\rowcolor{orange!15}
OpenResearcher 
& 26.3 & 54.8 & 14.8$^{\dagger}$ & 19.2$^{\dagger}$ & 19.6$^{\dagger}$ & 64.1 & 35.4$^{\dagger}$ & 61.3$^{\dagger}$ \\
\rowcolor{orange!15}
DR Tulu 
& - & - & 1.6$^{\dagger}$ & - & - & - & 45.5 & - \\

\midrule
\rowcolor{cyan!15}
\ours{}-30B 
& 37.0 & 48.2 & \underline{28.6} & 54.2 & 24.6 & 69.0 & 45.3 & \textbf{74.1} \\
\rowcolor{cyan!15}
\ours{}-35B 
& 45.5 / \underline{64.6}$^{*}$ & 61.0 / 69.5$^{*}$ & \textbf{30.7} & \underline{60.6} & 37.2 & \textbf{80.8} & 48.2 & 68.2 \\
\bottomrule
\end{tabular}%
}
\vspace{-1em}
\end{table}

\subsection{Results} \label{sec:exp:results}

\paragraph{\ourmodel{} establishes a new state of the art among open-weight agents and approaches proprietary agents.}
As shown in Table~\ref{tab:main_results},
across almost all evaluated benchmarks, \ourmodel\ consistently achieves new state-of-the-art performance among open-weight agents at around the 30B scale (e.g., Tongyi-DR, RedSearcher).
On several benchmarks, \ourmodel{} even matches or slightly surpasses closed-source agents, including OpenAI-DR on DeepResearch Bench (48.2\% vs.\ 47.0\%) and Mind2Web 2 (30.7\% vs.\ 28.0\%), as well as GPT-5 on GAIA (80.8\% vs.\ 76.4\%). 
These results substantially narrow the gap between open-weight and proprietary deep research systems.
These results highlight the effectiveness of our training recipe.

\paragraph{Existing open-weight agents fall short on out-of-distribution tasks, while \ours\ recipe leads to overall best performance.}
For a fair comparison with open-weight models at exactly the 30B scale, we further apply the \ours{} recipe to Qwen3-30B-A3B~\citep{yang2025qwen3}, resulting in our \ourtb{} model. 
We directly compare \ourtb{} with Tongyi-DR and OpenResearcher.
At the same parameter scale, our model achieves the best performance on 4 out of 8 benchmarks, including tasks that require strong \capTwo{} and \capThree{} capabilities, such as Mind2Web 2 and DeepResearch Bench. 
On the remaining benchmarks, Tongyi-DR performs best on BrowseComp, HLE, and GAIA, which rely heavily on the \capOne{} ability, a capability explicitly supported by Tongyi-DR's single-answer synthetic data.
In contrast, OpenResearcher performs best on BrowseComp-Plus, a fully offline benchmark that closely matches its data synthesis recipe. 
These results suggest that the capabilities exhibited by a deep research agent are shaped by its data synthesis recipe, and our recipe provides the best coverage of these capabilities.

\subsection{Insights from Controlled Studies}
\begin{figure*}[t]
\centering
\resizebox{\linewidth}{!}{
\begin{tikzpicture}
\begin{groupplot}[
  group style={group size=4 by 2, horizontal sep=0.9cm, vertical sep=0.8cm},
  bench line,
]
\nextgroupplot[xlabel={BrowseComp}, ymin=36, ymax=48, ytick={36,39,42,45,48}]
\addplot[mark=*, mark size=2pt, color=ColStage, thick, mark options={fill=ColStage}]
  coordinates {(Vanilla,38.4) (SFT,45.1) (+MT,45.5) (+RL,45.5)};
\lbl[anchor=north west, xshift=-9pt, yshift=14pt]{Vanilla}{38.4}{38.4}
\lbl[anchor=south east, xshift=8pt, yshift=2pt]{SFT}{45.1}{45.1}
\lbl[anchor=south west, xshift=-8pt, yshift=1pt]{+MT}{45.5}{45.5}
\lbl[anchor=north, xshift=-1pt, yshift=12pt]{+RL}{45.5}{45.5}

\nextgroupplot[xlabel={Mind2Web 2}, ymin=12, ymax=32, ytick={12,17,22,27,32}]
\addplot[mark=*, mark size=2pt, color=ColStage, thick, mark options={fill=ColStage}]
  coordinates {(Vanilla,15.1) (SFT,26.5) (+MT,29.9)  (+RL,30.7)};
\lbl[anchor=south west, xshift=1pt, yshift=-7pt]{Vanilla}{15.1}{15.1}
\lbl[anchor=south east, xshift=12pt, yshift=-13pt]{SFT}{26.5}{26.5}
\lbl[anchor=south west, xshift=-6pt, yshift=-13pt]{+MT}{29.9}{29.9}
\lbl[anchor=north, xshift=0.5pt, yshift=-2pt]{+RL}{30.7}{30.7}

\nextgroupplot[xlabel={HLE}, ymin=30, ymax=42, ytick={30,33,36,39,42}]
\addplot[mark=*, mark size=2pt, color=ColStage, thick, mark options={fill=ColStage}]
  coordinates {(Vanilla,32.3) (SFT,39.49) (+MT,39.74) (+RL,37.9)};
\lbl[anchor=north west, xshift=-10pt, yshift=13pt]{Vanilla}{32.3}{32.3}
\lbl[anchor=south east, xshift=8pt,yshift=1pt]{SFT}{39.49}{39.5}
\lbl[anchor=south west, xshift=-10pt, yshift=0pt]{+MT}{39.74}{39.7}
\lbl[anchor=north, xshift=1pt, yshift=13pt]{+RL}{37.9}{37.9}

\nextgroupplot[xlabel={DeepResearch Bench}, ymin=34, ymax=50, ytick={34,38,42,46,50}]
\addplot[mark=*, mark size=2pt, color=ColStage, thick, mark options={fill=ColStage}]
coordinates {(Vanilla,44.05) (SFT,36.35) (+MT,39.72) (+RL,48.15)};
\lbl[anchor=south west, xshift=-8pt, yshift=1pt]{Vanilla}{44.05}{44.1}
\lbl[anchor=north, yshift=13pt]{SFT}{36.35}{36.4}
\lbl[anchor=south east, xshift=6pt, yshift=2pt]{+MT}{39.72}{39.7}
\lbl[anchor=south east, xshift=-1pt, yshift=-3.5pt]{+RL}{48.15}{48.2}

\nextgroupplot[xlabel={BrowseComp Plus}, ymin=57, ymax=61.5, ytick={57,58,59,60,61}]
\addplot[mark=*, mark size=2pt, color=ColStage, thick, mark options={fill=ColStage}]
  coordinates {(Vanilla,58.5) (SFT,57.9) (+MT,58.6) (+RL,61.0)};
\lbl[anchor=south west, xshift=-10pt, yshift=-14pt]{Vanilla}{58.5}{58.5}
\lbl[anchor=north, yshift=-0.5pt]{SFT}{57.87}{57.9}
\lbl[anchor=south, xshift=1pt, yshift=-13pt]{+MT}{58.6}{58.6}
\lbl[anchor=north, xshift=1pt, yshift=-2pt]{+RL}{61.0}{61.0}

\nextgroupplot[xlabel={WideSearch}, ymin=38, ymax=70, ytick={38,46,54,62,70}]
\addplot[mark=*, mark size=2pt, color=ColStage, thick, mark options={fill=ColStage}]
  coordinates {(Vanilla,43.8) (SFT,61.1) (+MT,62.5) (+RL,64.5)};
\lbl[anchor=north west, xshift=-2pt, yshift=0pt]{Vanilla}{43.8}{43.8}
\lbl[anchor=south east, xshift=11pt, yshift=-13pt]{SFT}{61.1}{61.1}
\lbl[anchor=north, xshift=1pt, yshift=-2pt]{+MT}{62.5}{62.5}
\lbl[anchor=north, xshift=1pt, yshift=-2pt]{+RL}{64.465}{64.5}

\nextgroupplot[xlabel={GAIA}, ymin=70, ymax=86, ytick={70,74,78,82,86}]
\addplot[mark=*, mark size=2pt, color=ColStage, thick, mark options={fill=ColStage}]
  coordinates {(Vanilla,72.8) (SFT,83.5) (+MT,83.17) (+RL,80.8)};
\lbl[anchor=north west, xshift=-10pt, yshift=14pt]{Vanilla}{72.8}{72.8}
\lbl[anchor=south east, xshift=9pt, yshift=-1pt]{SFT}{83.5}{83.5}
\lbl[anchor=south west, xshift=-8pt, yshift=0pt]{+MT}{83.17}{83.2}
\lbl[anchor=south, xshift=1pt, yshift=2pt]{+RL}{80.8}{80.8}

\nextgroupplot[xlabel={LiveResearchBench}, ymin=64, ymax=68.5, ytick={64,65,66,67,68}]
\addplot[mark=*, mark size=2pt, color=ColStage, thick, mark options={fill=ColStage}]
coordinates {(Vanilla,65.02) (SFT,64.69) (+MT,65.47) (+RL,68.2)};
\lbl[anchor=south west, xshift=-9pt, yshift=1pt]{Vanilla}{65.02}{65.0}
\lbl[anchor=south, xshift=0.5pt, yshift=4pt]{SFT}{64.69}{64.7} 
\lbl[anchor=south west, yshift=2pt, xshift=-12pt]{+MT}{65.47}{65.5}
\lbl[anchor=east, yshift=1pt]{+RL}{68.15}{68.2}

\end{groupplot}
\end{tikzpicture}}
\vspace{-1.5em}
\caption{Ablation study on the effect of training stages. Score progression of \ourmodel\ across three training stages. All results are based on agents using the context condenser.
}
\vspace{-1em}
\label{fig:sft_mt_rl_ablation}
\end{figure*}
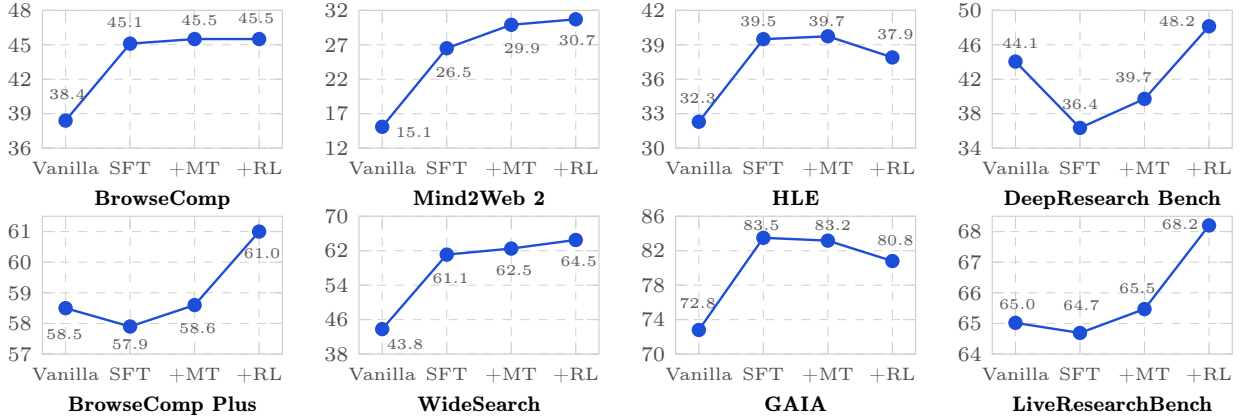

\paragraph{Best Performance Obtained by Combining MT, SFT, and RL.}
To better understand the impact of our training stages and training data on final performance, we systematically analyze \ourmodel{} under different stages. 
We consider four model variants: 
\textit{1) Vanilla}, the base checkpoint identical to Qwen3.5-35B-A3B;
\textit{2) SFT}, obtained by applying SFT to the Vanilla model; 
\textit{3) MT+SFT}, obtained by first performing MT on the Vanilla model followed by SFT; 
\textit{4) MT+SFT+RL}, our full training recipe, which further applies RL on top of MT+SFT. 
Figure~\ref{fig:sft_mt_rl_ablation} presents the results. 

We observe that the effect of each training stage is nuanced across benchmarks.
Simple SFT improves most objective benchmarks, but degrades open-ended performance relative to the Vanilla baseline.
The only objective benchmark that SFT hurts is BC-Plus, because it allows only a single \texttt{search} tool during evaluation, while heavy SFT makes the model more likely to call disallowed tools due to overfitting to the training-time tool-use pattern.
MT further improves performance on top of SFT overall, suggesting that the benefit of our auxiliary tasks generalizes across benchmarks.
Interestingly, RL substantially improves performance on open-ended tasks, but slightly sacrifices performance on HLE and GAIA.
We hypothesize that HLE and GAIA require not only search, but strong expert-level reasoning and in-depth analysis.
Since our RL objective goals are primarily optimized for deep research behaviors, including report synthesis like readability-oriented responses, this specialization may partially weaken the model's general reasoning capability, a phenomenon similar to the \textit{alignment tax}~\citep{ouyang2022traininglanguagemodelsfollow}.
Overall, \textit{MT+SFT+RL} performs best among the compared variants, demonstrating that general-purpose deep research benefits from all three training stages.

\definecolor{Col35B}{RGB}{58,85,227}
\begin{figure*}[t]
\begin{center}
\small
\fcolorbox{black!15}{blue!2}{
\hspace{6pt}
\begin{tabular}{cccc}
\legendsq{Col2B}\,\ours{}-2B-SFT &
\legendsq{Col4B}\,\ours{}-4B-SFT &
\legendsq{Col9B}\,\ours{}-9B-SFT &
\legendsq{Col35B}\,\ours{}-35B-SFT
\end{tabular}
\hspace{6pt}
}
\end{center}

\centerline{\resizebox{\linewidth}{!}{
\begin{tikzpicture}

\begin{groupplot}[
  group style={
    group size=4 by 2,
    horizontal sep=0.9cm,
    vertical sep=0.7cm
  },
  bench bar,
]

\nextgroupplot[
xlabel={BrowseComp},
ymin=20, ymax=52,
ytick={20,30,40,50}
]
\addplot[fill=Col2B,  draw=Col2B!60] coordinates {(A,28)};
\addplot[fill=Col4B,  draw=Col4B!60] coordinates {(B,40)};
\addplot[fill=Col9B,  draw=Col9B!60] coordinates {(C,45.4)};
\addplot[fill=Col35B, draw=Col35B!60] coordinates {(D,45.1)};

\nextgroupplot[
xlabel={Mind2Web 2},
ymin=0, ymax=32,
ytick={0,10,20,30}
]
\addplot[fill=Col2B,  draw=Col2B!60] coordinates {(A,8.8)};
\addplot[fill=Col4B,  draw=Col4B!60] coordinates {(B,24.3)};
\addplot[fill=Col9B,  draw=Col9B!60] coordinates {(C,24.4)};
\addplot[fill=Col35B, draw=Col35B!60] coordinates {(D,26.5)};

\nextgroupplot[
xlabel={HLE},
ymin=25, ymax=42,
ytick={25,30,35,40}
]
\addplot[fill=Col2B,  draw=Col2B!60] coordinates {(A,30.3)};
\addplot[fill=Col4B,  draw=Col4B!60] coordinates {(B,36.2)};
\addplot[fill=Col9B,  draw=Col9B!60] coordinates {(C,36.9)};
\addplot[fill=Col35B, draw=Col35B!60] coordinates {(D,39.5)};

\nextgroupplot[
xlabel={DeepResearch Bench},
ymin=15, ymax=42,
ytick={15,20,25,30,35,40}
]
\addplot[fill=Col2B,  draw=Col2B!60] coordinates {(A,21.0)};
\addplot[fill=Col4B,  draw=Col4B!60] coordinates {(B,22.0)};
\addplot[fill=Col9B,  draw=Col9B!60] coordinates {(C,32.6)};
\addplot[fill=Col35B, draw=Col35B!60] coordinates {(D,36.4)};

\nextgroupplot[
xlabel={BrowseComp-Plus},
ymin=45, ymax=62,
ytick={45,50,55,60}
]
\addplot[fill=Col2B,  draw=Col2B!60] coordinates {(A,52.6)};
\addplot[fill=Col4B,  draw=Col4B!60] coordinates {(B,52.1)};
\addplot[fill=Col9B,  draw=Col9B!60] coordinates {(C,55.6)};
\addplot[fill=Col35B, draw=Col35B!60] coordinates {(D,57.9)};

\nextgroupplot[
xlabel={WideSearch},
ymin=35, ymax=67,
ytick={35,45,55,65}
]
\addplot[fill=Col2B,  draw=Col2B!60] coordinates {(A,40.9)};
\addplot[fill=Col4B,  draw=Col4B!60] coordinates {(B,55.0)};
\addplot[fill=Col9B,  draw=Col9B!60] coordinates {(C,58.5)};
\addplot[fill=Col35B, draw=Col35B!60] coordinates {(D,61.1)};

\nextgroupplot[
xlabel={GAIA},
ymin=70, ymax=87,
ytick={70,75,80,85}
]
\addplot[fill=Col2B,  draw=Col2B!60] coordinates {(A,72.8)};
\addplot[fill=Col4B,  draw=Col4B!60] coordinates {(B,77.7)};
\addplot[fill=Col9B,  draw=Col9B!60] coordinates {(C,78.6)};
\addplot[fill=Col35B, draw=Col35B!60] coordinates {(D,83.5)};

\nextgroupplot[
xlabel={LiveResearchBench},
ymin=55, ymax=67,
ytick={55,60,65}
]
\addplot[fill=Col2B,  draw=Col2B!60] coordinates {(A,57.4)};
\addplot[fill=Col4B,  draw=Col4B!60] coordinates {(B,62.1)};
\addplot[fill=Col9B,  draw=Col9B!60] coordinates {(C,63.5)};
\addplot[fill=Col35B, draw=Col35B!60] coordinates {(D,64.7)};

\end{groupplot}
\end{tikzpicture}}}
\vspace{-1em}
\caption{
Benchmark performance across different \ours{} model sizes.
For BrowseComp, HLE, and BrowseComp-Plus, we evaluate on subsets.
For GAIA, we report results on GAIA-Text-103.
}
\label{fig:smaller_versions}

\end{figure*}
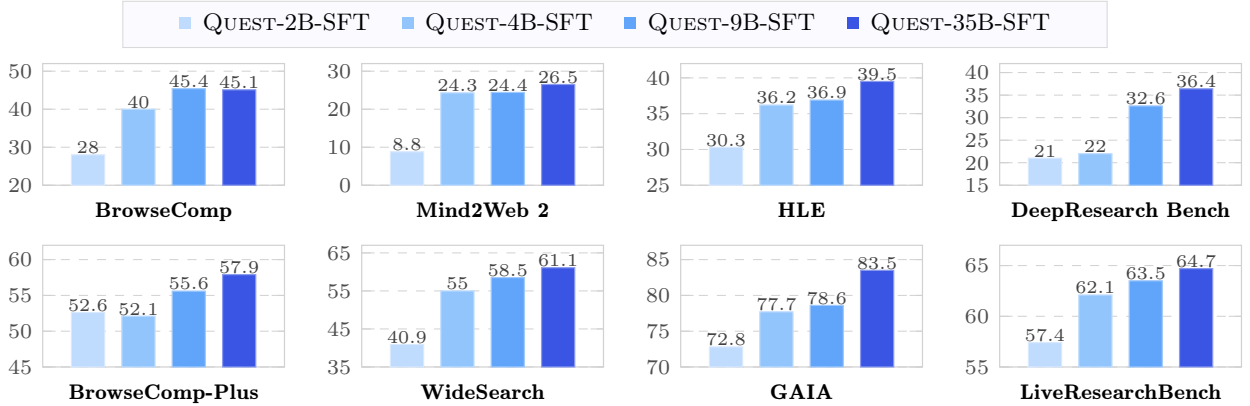

\paragraph{Small LLMs Can be Strong Deep Research Agents.}
Developing capable agents based on smaller models is practically meaningful, especially for compute-constrained scenarios or high-stakes domains such as healthcare, where hosting agents locally is desired due to privacy concerns.
To study this, we conduct SFT on the 2B, 4B, and 9B versions of Qwen3.5 \citep{qwen3.5} using the same training data and inference configurations as our previous experiments, which allows us to study how performance changes as model size scales.

Surprisingly, as shown in Figure~\ref{fig:smaller_versions}, %
downscaling model size does not lead to substantial degradation on most benchmarks, although overall performance generally improves with larger LLMs.
Even a 2B model achieves competitive performance on \capOne\ benchmarks such as HLE and GAIA, matching or even outperforming models like OpenAI-DeepResearch and o3~\citep{openai2025o4mini, openai2025deepresearch} (e.g., o3 scores 24.9\% on HLE and 70.5\% on GAIA).
This highlights the robustness of our synthetic data across model scales, 
as SFT alone yields strong performance even with a 2B model.
However, this advantage does not fully transfer to benchmarks that heavily require \capThree, such as DeepResearch Bench and LiveResearchBench, where the 2B model lags far behind.
Overall, our findings suggest a promising direction for deploying small LLMs for \capOne\ search tasks, while also highlighting the challenge of extending their capabilities to more open-ended settings.

\section{Unsuccessful Attempts}
In this section, we share some of our unsuccessful attempts during the preliminary experimental stage below, with the hope that these empirical insights will inform and benefit future efforts.
\subsection{Search Result Prediction in Mid-training}
Prior work~\citep{zhang2025ee} has shown that search result prediction can be beneficial for search scenarios.
Motivated by this, we incorporate a similar objective into our mid-training pipeline.
Specifically, following~\citet{zhang2025ee}, we summarize retrieved search results and formulate the task as next-token prediction over the summarized content.
When used in isolation, this objective yields performance improvements, especially on some search-heavy tasks, like BrowseComp.
However, adding it to our current mid-training objective mixture consistently hurts overall performance.
We hypothesize that this interference arises from redundancy with our context-related objectives.
In particular, the \textit{Context Summarization} task also requires the model to condense search results into a structured Context State, which may overlap with the functionality of search result prediction.
This redundancy could lead to conflicting learning signals during joint training.
We leave a more principled integration of these objectives to future work.

\subsection{Rubric-Based Error Identification in Mid-training}
Given a generated answer, a rubric tree can provide fine-grained error analysis, which suggests a potential training signal for teaching the model to identify its own mistakes and improve response quality.
Motivated by this, we introduce a mid-training task where the model is asked to predict potential issues in a given answer.
However, in our preliminary experiments, this objective yields only marginal improvements.
We hypothesize that this limitation stems from the lack of access to external evidence during this stage.
As a result, the model can only identify superficial or commonsense errors, while failing to detect more complex factual or evidence-dependent issues, which are relatively rare in generated responses.
This suggests that rubric-based error prediction alone is insufficient as a standalone training signal without grounding in external knowledge.

\subsection{Direct Preference Optimization (DPO) for Report Comparison}
To improve report generation on open-ended tasks, we explore a direct preference optimization objective based on pairwise report comparison.
Given two candidate reports for the same task, we construct preference pairs using rubric-tree scores and train the policy to favor higher-quality outputs, following the DPO paradigm~\citep{rafailov2023direct}.
However, in our experiments, this approach does not yield improvements.
Constructing reliable preference pairs incurs additional overhead, and training tends to be unstable and prone to overfitting, resulting in limited gains in final report quality.
We hypothesize that this limitation arises from the inherently high variance and ambiguity in long-form report comparison.
Unlike short responses, reports often differ along multiple dimensions (e.g., coverage, organization, and evidence use), making it difficult to derive reliable pairwise preferences.
Moreover, small differences in rubric scores may not correspond to meaningful qualitative distinctions, further weakening the training signal.

\subsection{Evaluation for Open-ended Tasks}
To provide reliable reward signals for open-ended task evaluation, we explored two scoring strategies before arriving at our final design.
\begin{enumerate}
    \item \textbf{Pointwise Scoring.}
    We first adopted a pointwise scoring method, where each rubric criterion is assessed on a three-tier scale: \textit{Not Satisfied} (0), \textit{Partially Satisfied} (0.5), and \textit{Satisfied} (1).
    However, this scheme suffered from severe score inflation: the score reached approximately 1 in $\sim50\%$ of cases, which makes it extremely difficult to distinguish between good and poor responses.
    We hypothesize this is due to the absence of a comparative reference: without a baseline, the judge model exhibits a strong high-score bias to favor the user.

    \item \textbf{Pairwise Win/Tie/Lose.}
    To introduce a comparative reference, we generated a report for each query using a teacher model and switched to a ternary pairwise scoring rule with scores $\{0, 0.5, 1\}$ corresponding to loss, tie, and win, respectively.
    However, since our model's early performance was consistently below that of the teacher, the judge assigned ``lose'' in nearly all cases, collapsing scores toward zero and rendering the signal unusable for both SFT filtering and RL training.
\end{enumerate}
\noindent
Both failure modes motivated our final pairwise scoring design described in Section~\ref{sec:data:obj}.

\section{Conclusion}
\ours{} is an open family of general-purpose deep research agents capable of
solving long-horizon tasks that require \capOne{}, \capThree{}, and \capTwo{}.
The largest model, \ourmodel, achieves the best overall performance among recent open-weight agents and approaches or even surpasses closed-source agents across eight benchmarks.
Beyond the models themselves, the central contribution is an open, reproducible recipe: rubric-tree-based data synthesis, 
structured context management, efficient training infrastructure, and a staged training pipeline combining mid-training, supervised finetuning, and RL. 
We hope this work provides a strong open foundation for future deep research agents.

\section{Acknowledgments}
The authors thank colleagues from the OSU NLP group for their constructive feedback. 
This research was supported in part by NSF CAREER Awards \#1942980 and \#2443149, the Alfred P. Sloan Research Fellowship, and supported by Amazon, Cisco, and Intuit.
We also acknowledge the computational resources provided by the Ohio Supercomputer Center \citep{ohio_supercomputer_center_1987}.
The views and conclusions contained herein are those of the authors and should not be interpreted as representing the official policies, either expressed or implied, of the U.S. Government. 
The U.S. Government is authorized to reproduce and distribute reprints for Government purposes, notwithstanding any copyright notice herein.

\clearpage
\bibliographystyle{plainnat}
\bibliography{sample}

\clearpage
\appendix
\section{Data Construction Details}
\label{app:data}
Below, we describe how our data are processed.

\subsection{Task Synthesis}
We adopt a strict task filtering strategy to ensure that all the generated tasks are high-quality and the evaluation trees are reliable.
This strict filtering policy leads to a low retention rate.
We report the number of remaining tasks at each stage in Table~\ref{tab:objective-filtering}.

\begin{table}[h]
\centering
\small
\caption{Number of retained tasks after each filtering stage.}
\begin{tabular}{l r}
\toprule
\textbf{Filtering Stage} & \textbf{Number of Tasks} \\
\midrule
\multicolumn{2}{l}{\textbf{Objective Tasks}} \\
\quad Initial generated objective tasks & 17,000 \\
\quad After rubric refinement & 8,737 \\
\quad After rubric tree verification & 6,230 \\
\quad After removing tasks with erroneous Python scripts & 5,934 \\
\midrule
\multicolumn{2}{l}{\textbf{Open-ended Tasks}} \\
\quad Initial generated open-ended tasks & 3,000 \\
\quad After removing tasks with invalid format & 2,856 \\
\quad After removing unqualified tasks & 2,498 \\
\quad After removing tasks with low reference report scores & 2,227 \\
\bottomrule
\end{tabular}
\label{tab:objective-filtering}
\end{table}

\subsection{Manual Examination of the Python Evaluation Scripts for Objective Tasks}
\label{app:python-script-check}
To examine the correctness of the Python evaluation scripts generated by \gsyn, we sample 50 tasks from our datasets and employ four graduate students majoring in computer science as human annotators for manual evaluation. 
The annotators are given the task description and the rubric tree, along with detailed instructions covering our assessment objectives, the organizational structure of the rubric tree, evaluation strategies, and core evaluation toolkit functionalities (such as Extractor and Verifier functions and rubric tree management utilities). 
We also provide examples of common mistakes and tips to help annotators identify potential errors.
Then they are required to verify whether the Python evaluation script aligns with the rubric tree and assess the program’s executability.

Out of the 50 sampled tasks, only 2 cases are found to suffer from non-executable script issues, such as runtime exceptions and logical execution breakdowns that prevent the evaluation pipeline from running normally.

In terms of rubric-related errors in the generated scripts, we identify 6 scripts that contain such errors, and summarize three typical problematic patterns. First, some scripts introduce redundant rubric nodes that are never required or mentioned in the original task description, adding unnecessary evaluation constraints irrelevant to the assessment goal. 
Second, some scripts perform verification merely based on the literal content of the final answer, rather than strictly verifying the authenticity of the content in accordance with the evaluation criteria. 
Third, some scripts fail to explicitly assign a negative judgment when the target answer contained no cited URLs, lacking clear judgment criteria for this scenario.

Despite imperfections, most generated scripts can accurately interpret task requirements, faithfully implement rubric rules, and run without errors, effectively supporting our automated evaluation at scale.

\subsection{Human Audit of the Final Training Data}
\label{app:human-audit}
To complement the automated filtering and the script examination above, we conduct a component-wise human audit of the final SFT dataset.
Before annotation, all annotators complete a calibration exercise on a small shared subset of examples to align the annotation criteria, after which we verify annotation consistency; the resulting agreement exceeds 95\% across all evaluated criteria.
All audited examples are randomly sampled from the final SFT dataset.

\textbf{Objective tasks.}
Ten annotators inspect 100 objective tasks.
The audit is fine-grained: a single example can contain more than 100 rubric nodes and over 2,000 lines of evaluator code, requiring more than one hour to inspect.
For each example, annotators assess
(D1) whether every rubric node is valid, relevant, and free of hard-coded answers;
(D2) whether the question is clear, solvable, and faithful to the rubric tree;
(D3) whether the executable evaluator implements the rubric tree and the task requirements; and
(D4) whether the final answer of the retained trajectory is correct.

\textbf{Open-ended tasks.}
Five annotators inspect 100 open-ended tasks, assessing
(D1) whether the task is clear and suitable for an evidence-backed long-form answer;
(D2) whether the rubric is faithful, factual, relevant, non-redundant, and covers the key aspects of the task; and
(D3) whether the final report of the retained trajectory is correct and comprehensive, with high-quality citations and coherent writing.

\begin{table}[h]
\centering
\small
\caption{Human audit of 100 objective and 100 open-ended tasks randomly sampled from the final SFT dataset.}
\label{tab:human-audit}
\begin{tabular}{l r}
\toprule
\textbf{Criterion} & \textbf{Pass Rate} \\
\midrule
\multicolumn{2}{l}{\textbf{Objective Tasks}} \\
\quad Rubric-tree quality (D1) & 99\% \\
\quad Question quality (D2) & 99\% \\
\quad Evaluator alignment (D3) & 97\% \\
\quad Trajectory correctness (D4) & 100\% \\
\midrule
\multicolumn{2}{l}{\textbf{Open-ended Tasks}} \\
\quad Task quality (D1) & 98\% \\
\quad Rubric quality (D2) & 96\% \\
\quad Trajectory correctness (D3) & 98\% \\
\bottomrule
\end{tabular}
\end{table}

As shown in Table~\ref{tab:human-audit}, all pass rates exceed 95\%.
For objective tasks, the errors comprise one hard-coded rubric node (D1), one question that omits a constraint specified by its rubric tree (D2), and three evaluator-alignment failures (D3): the same hard-coding issue and two evaluators that add constraints absent from their rubric trees.
Since the natural-language question is generated from the rubric tree rather than the reverse, the rubric tree defines the task intent; the single D2 failure is the only case in which question generation drifted from it.
The evaluator alignment rate is higher than that in Appendix~\ref{app:python-script-check} because rejection sampling removes many low-quality intermediate examples before the final SFT dataset is formed.
For open-ended tasks, the failures comprise two underspecified tasks (D1); four rubric issues (D2), namely one missing aspect, two partially redundant criteria, and one unnecessary requirement; and two trajectory issues (D3), namely one report that omits key criteria and one that contains repetitive paragraphs.

\textbf{Verifier accuracy.}
We further compare 110 judgments produced by the generated Python evaluators with human annotations.
The evaluators agree with humans in 108 of the 110 cases (98.2\%).
Both errors are false acceptances, i.e., incorrect answers judged as correct, and no correct answer is rejected.
Since each evaluator decomposes the task into atomic rubric-node checks, an error at a single node affects only the corresponding branch, while the remaining branches still provide valid partial-credit signals.

\subsection{Mid-Training}
\label{app:data-mt}
\paragraph{Context Summarization.}
The context summarization data is collected from the distillation process used for SFT data preparation and from several preliminary RL experiments.
We filter out low-quality instances caused by API errors or other implementation bugs.
To avoid data leakage, we do not include any context summarization data from benchmark evaluations.
This results in a final set of 300K context summarization instances.

\paragraph{Relevant Information Extraction.}
The relevant information extraction data is obtained from our Visit Tool cache, which stores the raw HTML content of all visited webpages, along with their corresponding extraction goals and extracted content.
Initially, this cache contains 4 million raw triplets of the form (raw HTML content, goal, extracted content).
To improve training efficiency and reduce redundancy, we filter repetitive instances, such as cases where multiple similar extraction goals correspond to the same webpage.
Specifically, we remove redundant instances by grouping entries from the same webpage and filtering similar extraction goals using Jaccard similarity with a threshold of 0.1.

\subsection{Supervised Fine-tuning}

For rejection sampling, we roll out each task five times.
For reflection-based retry, we roll out each failed trajectory three additional times.
All the maximum tool calls are set to 100 turns. 
And the context threshold is set to 16K.

\subsection{Reinforcement Learning}
Tasks that cannot be successfully completed by \gtraj{} in any rollout are reserved for reinforcement learning, resulting in 864 objective RL tasks.
For open-ended tasks, we manually split the data into SFT and RL sets.

\section{Session-level Training}
\label{app:session-level-training}
\paragraph{Memory Efficiency.}
Treating an entire trajectory as a single training sequence requires the full context to reside in GPU memory simultaneously, which is infeasible for long-horizon trajectories.
Decomposing trajectories into shorter sessions substantially reduces peak memory usage, allowing training to scale to longer trajectories without being limited by GPU memory.
In practice, this allows us to train on trajectories exceeding 200K tokens using only 16 H100 GPUs, which would be infeasible under standard trajectory-level training.

\paragraph{Unbounded Context Extrapolation at Inference Time.}
Session-level training exposes the model to variable-length inputs under a structured decomposition, effectively decoupling per-session training context length from overall trajectory length. 
Combined with our context management mechanism, this enables the agent to extrapolate to arbitrarily long research sessions at inference time, far beyond the context lengths seen during training and even beyond the model's native context window. 
For example, although \ours{} is trained on trajectories of at most 100 turns, it is able to generalize to interactions of more than 200 turns at inference time.

\section{Context Management}
\subsection{Why Context Condenser instead of Discard-All or Keep-Last-N}

A natural alternative to a structured context condenser is to simply discard old context once the window fills, either by clearing everything (\textit{discard-all}) or by retaining only the most recent $N$ turns (\textit{keep-last-N})~\citep{glm5team2026glm5vibecodingagentic}.
While simple to implement, both strategies suffer from a fundamental limitation: they treat all historical information as equally disposable, with no mechanism to distinguish verified facts from speculative claims, or completed sub-goals from pending ones.

Discard-all resets the agent's epistemic state entirely, forcing it to rediscover facts it has already verified and revisit sources it has already read.
Keep-last-N mitigates this to some extent by preserving recent context, but remains blind to information that falls outside the retention window, even if that information is critical to the task.
In both cases, the agent loses not just raw content, but structured knowledge about \textit{what it knows}, \textit{what it doubts}, and \textit{what it still needs to find}.

While discard-all and keep-last-N may suffice at inference time for benchmarks with short horizons, they are fundamentally inadequate as a training paradigm. 
A reliable training signal requires the agent to maintain a coherent, consistent view of its research state across the entire trajectory, particularly under constrained context budgets (e.g., 32K), where naive truncation would corrupt the very supervision signal we rely on.
The Context Condenser addresses this by compressing history into a structured epistemic snapshot rather than a raw truncation: verified facts are carried forward explicitly, pending actions are preserved as concrete follow-up directives, and contradicted claims are flagged rather than silently dropped.
This ensures that no verified knowledge is lost across compression boundaries, and both training trajectories and inference rollouts remain coherent regardless of task length.

\section{Model Configuration}
\label{app:model-config}
\begin{table*}[t]

\caption{Models used across \ours{} inference, task-generation, SFT, and RL stages.}

\label{tab:default-models}

\centering

\small

\begin{tabularx}{\linewidth}{l l X}

\toprule

\textbf{Stage} & \textbf{Use} & \textbf{Default Model} \\

\midrule

Inference & Visit Summary Model & GPT-5-mini \\

Inference & Context Condenser & GPT-5-mini \\

\midrule

Objective Task Generation & Main Generation Model & Claude Sonnet 4.5 \\

Objective Task Generation & Visit Summary Model & GPT-5-mini \\

Objective Task Generation & Rubric Refinement Model & GPT-5.2 \\

Objective Task Generation & Rubric Verification Model & Claude Sonnet 4.5 \\

\midrule

Open-ended Task Generation & Main Generation Model & Claude Sonnet 4.5 \\

Open-ended Task Generation & Visit Summary Model & GPT-5-mini \\

Open-ended Reference Generation & Reference Answer Generation Model & Claude Sonnet 4.5 \\

\midrule

Distillation Model & Trajectory Distillation for SFT & Tongyi-DeepResearch \\

\midrule

RL Training & Judge Model & Qwen3-4B \\

RL Training & Judge Fallback Model  & DeepSeek V3.2 \\

RL Training & Visit Summary Model & GPT-5-mini \\

RL Training & Visit Summary Fallback Model & DeepSeek V3.2 \\

RL Training & Context Condenser & GPT-5-mini \\

RL Training & Context Condenser Fallback Model & DeepSeek V3.2 \\

RL Training & Fact-checking Model & GPT-5-mini \\

RL Training & Fact-checking Fallback Model & DeepSeek V3.2 \\

\bottomrule

\end{tabularx}

\end{table*}
In Table~\ref{tab:default-models}, we provide the list of models used across \ours{} inference, task generation, SFT, and RL stages.
To ensure stable RL training, we use fallback models (DeepSeek V3.2) for the context condenser and visit summarization modules, preventing training from being interrupted by RPM (requests-per-minute) limits.
The fallback trigger rate is low, around 3\% in our statistics, but without this mechanism, RL training becomes unstable.

\section{Training Details}
\label{app:training-details}
\subsection{Mid-Training}
We conduct mid-training for one epoch with full-parameter optimization. Following our Megatron-LlamaFactory configuration, we use BF16 training, a learning rate of $3\times10^{-6}$ with a constant scheduler, a maximum context length of 24,576 tokens, per-device batch size 2, and gradient accumulation steps 8. Mid-training is performed on 32 H100 GPUs and takes approximately 5 days.

\subsection{Supervised Fine-Tuning}
After mid-training, we further perform supervised fine-tuning with full-parameter optimization. We use BF16 training, a learning rate of $3\times10^{-6}$ with a constant scheduler, a maximum sequence length of 32{,}768 tokens, per-device batch size 1, and gradient accumulation steps 8 for 3 epochs. 
Supervised fine-tuning is also conducted on 32 H100 GPUs and takes approximately 1 day.

\subsection{Reinforcement Learning}
After supervised fine-tuning, we further optimize the model with a fully asynchronous RL pipeline based on VERL.
We use GRPO as the advantage estimator, with actor learning rate $1\times10^{-6}$, PPO mini-batch size 16, and 8 responses sampled per prompt. 
The global batch size is 64, which is divided into four PPO mini-batches; the actor is updated on each mini-batch, while the updated policy parameters are synchronized to the rollout workers only after all four mini-batch updates are completed. 
The maximum prompt length is 24,000 tokens, the maximum response length is 12,288 tokens, and the maximum single-turn response length is 10,240 tokens. 
During rollout, we use a temperature of 1.0 and allow up to 100 tool-call turns.
The context threshold is set to 16K tokens.
In the default fully asynchronous setup, training and rollout are decoupled: the trainer uses 16 GPUs, while rollout uses 16 GPUs, for a total of 32 H100 GPUs. 
We run RL for 80 asynchronous training steps, which corresponds to 320 mini-batch-level optimization steps under our 4-step update schedule.

\section{Evaluation Subset}
\label{app:subset}
Due to the long-horizon nature of deep research tasks, evaluating on some large-scale benchmarks can be expensive in terms of both time and API cost (e.g., BrowseComp contains 1,266 instances and HLE-Text contains 2,158 instances).
To reduce evaluation cost, in the ablation study and analyses, we report results on randomly sampled subsets for these benchmarks.
For WideSearch, we use its English subset. 
All other subsets are sampled uniformly at random without manual filtering.
We provide the detailed statistics in Table~\ref{tab:subset}.

\section{Evaluation Details of the Vanilla Model}
\label{app:evaluation-qwen35}
In Qwen3.5's technical blog~\citep{qwen3.5}, Qwen Team reported their performance on the HLE (w/ tool) and Browsecomp benchmarks, employing a context management strategy.
Based on the limited information disclosed, we attempted to reproduce their results on these benchmarks.
Although we strictly followed their context management approach, adopted official tool-calling format, and utilized the same tool-set as \ours{}, we consistently obtained lower scores and were unable to replicate their reported performance. 
We hypothesize that this discrepancy may stem from unrevealed differences in prompts, tool specifications, implementation details (e.g., handling scenarios where websites are blocked or intercepted by anti-bot systems), or nuances in the context management strategy. 
Given the lack of specific details in their technical blog regarding the evaluation configuration, we opted to use the same prompt, tool-set and context management strategy as \ours{} in our evaluation. 
This approach ensures a fair and consistent baseline to better assess the effectiveness of each stage in our proposed training recipe.

\section{Additional Experiments}
\label{app:additional-exp}

\subsection{Human Evaluation of Open-Ended Reports}
\label{app:human-eval-report}
To complement benchmark and LLM-judge scores, we conduct a blind human evaluation of open-ended reports.
Four annotators rank the reports produced by OpenAI DeepResearch, Gemini DeepResearch, Tongyi DeepResearch, and \ours{} on 100 DeepResearch Bench tasks.
Model identities are hidden and the order of reports is randomized.
Rankings (1 = best, 4 = worst) jointly consider readability, insightfulness, comprehensiveness, and instruction following.
For the proprietary systems, we use the versions accessible at the time of this study, i.e., GPT-5.6 Sol Medium for OpenAI DeepResearch and Gemini 3.6 Flash for Gemini DeepResearch, since legacy models used in earlier studies (e.g., Gemini 2.5 Pro) are no longer available through current interfaces.
Tongyi DeepResearch is run locally with its official harness.

\begin{table}[h]
\centering
\small
\caption{Average human ranking of reports on 100 DeepResearch Bench tasks (lower is better).}
\label{tab:human-eval-report}
\begin{tabular}{l c c c c}
\toprule
 & \textbf{OpenAI-DR} & \textbf{Gemini-DR} & \textbf{Tongyi-DR} & \textbf{\ours{}} \\
 & (GPT-5.6 Sol Medium) & (Gemini 3.6 Flash) & & \\
\midrule
Average rank $\downarrow$ & 1.83 & 2.09 & 3.79 & 2.29 \\
\bottomrule
\end{tabular}
\end{table}

As shown in Table~\ref{tab:human-eval-report}, \ours{} substantially outperforms the open-weight Tongyi DeepResearch and trails Gemini DeepResearch by only 0.20 in average rank, although the continuously updated proprietary systems remain the strongest.

\subsection{Sensitivity to the Judge Model}
\label{app:judge-sensitivity}
The reward for open-ended tasks relies on a reference report and an LLM judge.
The reference report serves only as a calibration anchor to reduce cross-task score drift and score inflation: the candidate and reference reports are scored separately against the task-specific rubric nodes, so a candidate with a different structure or writing style can still outperform the reference.
In our preliminary evaluation, removing the reference anchor caused roughly half of the responses to receive scores close to 1, indicating poor score calibration.

\textbf{Open-ended scoring.}
We re-evaluate 45 open-ended tasks by replacing the original Qwen3-4B-Instruct judge with GPT-OSS-20B.
For each task, we rank the same three candidate responses under each judge and measure rank agreement with Kendall's $\tau$~\citep{kendall1938new}.
The mean Kendall's $\tau$ is 0.674.
Specifically, 28 of the 45 tasks (62.2\%) have identical rankings ($\tau=1$), 12 (26.7\%) differ by one pairwise inversion ($\tau=1/3$), and 5 (11.1\%) differ by two pairwise inversions ($\tau=-1/3$); no task exhibits a complete ranking reversal ($\tau=-1$).
Manual inspection shows that disagreements are concentrated on responses with close scores rather than on clearly implausible rankings.

\textbf{Citation checking.}
We sample 500 claim--URL pairs from the outputs of the three evaluated systems and re-annotate them with the same two judges.
The two judges agree on 91.8\% of the labels over the three classes (supported, unsupported, and unknown), and their system-level citation-support rates are highly correlated ($r=0.93$).
Most disagreements involve inaccessible webpages or claims only partially supported by the cited source.
Together with the fact that the fact-checking reward contributes at most 25\% of the total reward, these results indicate that judge choice affects some close cases but does not materially alter the comparative conclusions.

\section{Inference Prompts}
\label{app:inference-prompts}

This section shows the prompts that drive the \ours{} inference pipeline at runtime.

\paragraph{Agent System Prompt.}
The following figure illustrates the system prompt of our agent. 
It defines the assistant's role, declares the \texttt{search}, \texttt{visit}, \texttt{google\_scholar}, and \texttt{PythonInterpreter} tool schemas in function-call format, instructs the model how to consume the \texttt{prev\_state} JSON emitted by the condenser, and pins down the required output formats: \texttt{<tool\_call>\{...\}</tool\_call>} for tool invocations and \texttt{<answer>...</answer>} for the final reply. The literal current date is appended at runtime.

\begin{promptbox}
You are a deep research assistant. Your core function is to conduct thorough, multi-source investigations into any topic. You must handle both broad, open-domain inquiries and queries within specialized academic fields. For every request, synthesize information from credible, diverse sources to deliver a comprehensive, accurate, and objective response. When you have gathered sufficient information and are ready to provide the definitive response, you must enclose the entire final answer within <answer></answer> tags.

# Tools

You may call one or more functions to assist with the user query.

You are provided with function signatures within <tools></tools> XML tags:
<tools>
{"type": "function", "function": {"name": "search", "description": "Perform Google web searches then returns a string of the top search results. Accepts multiple queries.", "parameters": {"type": "object", "properties": {"query": {"type": "array", "items": {"type": "string", "description": "The search query."}, "minItems": 1, "description": "The list of search queries."}}, "required": ["query"]}}}
{"type": "function", "function": {"name": "visit", "description": "Visit webpage(s) and return the summary of the content.", "parameters": {"type": "object", "properties": {"url": {"type": "array", "items": {"type": "string"}, "description": "The URL(s) of the webpage(s) to visit. Can be a single URL or an array of URLs."}, "goal": {"type": "string", "description": "The specific information goal for visiting webpage(s)."}}, "required": ["url", "goal"]}}}
{"type": "function", "function": {"name": "google_scholar", "description": "Leverage Google Scholar to retrieve relevant information from academic publications. Accepts multiple queries. This tool will also return results from google search", "parameters": {"type": "object", "properties": {"query": {"type": "array", "items": {"type": "string", "description": "The search query."}, "minItems": 1, "description": "The list of search queries for Google Scholar."}}, "required": ["query"]}}}
{"type": "function", "function": {"name": "PythonInterpreter", "description": "Executes Python code in a sandboxed environment. To use this tool, you must follow this format:
1. The 'arguments' JSON object must be empty: {}.
2. The Python code to be executed must be placed immediately after the JSON block, enclosed within <code> and </code> tags.

IMPORTANT: Any output you want to see MUST be printed to standard output using the print() function.

Example of a correct call:
<tool_call>
{"name": "PythonInterpreter", "arguments": {}}
<code>
import numpy as np
# Your code here
print(f"The result is: {np.mean([1,2,3])}")
</code>
</tool_call>", "parameters": {"type": "object", "properties": {}}}}
</tools>

# Using prev_state (Research State Summary)

If you see a "RESEARCH STATE SUMMARY (prev_state)" section in the user message, it contains a compressed summary of previous research progress. Use it to:

1. **Avoid redundant work**: 
   - Check `search_queries` to avoid repeating searches that have already been executed.
   - Check `visited_sources` to avoid visiting URLs that have already been visited.

2. **Use verified information**:
   - Check `information_state.trusted` for facts that have been verified from visited sources. You can use these directly in your answer without re-searching or re-visiting.
   - Check `information_state.untrusted` for claims that have been contradicted or proven unreliable.

3. **Follow up on uncertain information**:
   - Check `information_state.uncertain` for claims that need more evidence. The `need` field specifies the exact next action (e.g., "visit <URL>" or "search <query>") to resolve the uncertainty.

IMPORTANT: Do NOT search for or visit information that is already in `prev_state`, unless it's insufficient to answer the user's question. Only in this case, you are encouraged to search for more information or even visit the same URL. Instead, use the information from `prev_state` directly, or follow the specific actions suggested in `information_state.uncertain.need` if more information is needed.

The final answer must exclude any information that remains uncertain or pending. All statements included must be fully verified.

For each function call, return a json object with function name and arguments within <tool_call></tool_call> XML tags:
<tool_call>
{"name": <function-name>, "arguments": <args-json-object>}
</tool_call>

Current date:
\end{promptbox}

\paragraph{Web-page Extractor Prompt.}
The following prompt is used inside the \texttt{Visit} tool. A lightweight LLM is given the raw page contents and the agent's stated goal, and returns a JSON with \texttt{rational}, \texttt{evidence}, and \texttt{summary} fields. This is what makes our visit tool return structured digest text rather than raw HTML.

\begin{promptbox}
Please process the following webpage content and user goal to extract relevant information:

## **Webpage Content** 
{webpage_content}

## **User Goal**
{goal}

## **Task Guidelines**
1. **Content Scanning for Rationale**: Locate the **specific sections/data** directly related to the user's goal within the webpage content
2. **Key Extraction for Evidence**: Identify and extract the **most relevant information** from the content, you never miss any important information, output the **full original context** of the content as far as possible, it can be more than three paragraphs.
3. **Summary Output for Summary**: Organize into a concise paragraph with logical flow, prioritizing clarity and judge the contribution of the information to the goal.

**Final Output Format using JSON format has "rational", "evidence", "summary" fields**
\end{promptbox}

\paragraph{Context Condenser Prompt.}
The following prompt is used for the Context Condenser when the agent's context exceeds the context threshold. 
It compresses the conversation into a strict JSON state with three buckets (\texttt{trusted}, \texttt{untrusted}, \texttt{uncertain}), tracks visited sources and prior search queries, and enforces deduplication and bucket-migration rules to prevent redundant searches.

\begin{promptbox}
You are a State Summarizer for a DeepResearch agent.
Your ONLY job is to maintain a compact, parseable, context-aware state JSON for context management.

Your primary objective is to prevent redundant search and redundant visit actions by extracting useful, answer-ready information from tool responses and preserving it in a structured state.

You will be given:
1) events: a chronological list of interaction events (user/assistant messages and tool calls/responses)
2) prev_state: the previous state JSON (may be empty or null)

You MUST output ONLY a single JSON object that conforms EXACTLY to the schema below.
No markdown, no extra text, no code fences, no explanations.

========================
OUTPUT JSON SCHEMA (STRICT)

{
  "version": "dr_state",
  "search_queries": [
    { "q": "string", "intent": "string" }
  ],
  "visited_sources": [
    { "url": "string", "note": "string" }
  ],
  "information_state": {
    "trusted": [
      { "id": "T1", "claim": "string", "sources": ["string"], "reason": "string" }
    ],
    "untrusted": [
      { "id": "U1", "claim": "string", "sources": ["string"], "reason": "string" }
    ],
    "uncertain": [
      { "id": "C1", "claim": "string", "sources": ["string"], "reason": "string", "need": "string" }
    ]
  }
}

========================
TRIGGER NOTE (IMPORTANT)

This summarizer is invoked automatically when CONTEXT_THRESHOLD is reached:
- The system invokes summarization when context tokens reach a threshold.
- Focus on extracting evidence, deduplicating tool usage, and making the state more actionable.

Note: Agent-initiated condenser tool calls are ignored for memory updates.
Only automatic CONTEXT_THRESHOLD triggers will update the memory state.

========================
CORE PRINCIPLE (CRITICAL)

Visited pages alone are NOT useful memory.

For every visit() tool_response, you MUST attempt to extract at least one
useful, concrete fact into information_state unless the page is irrelevant.

The goal is that the DeepResearch agent can rely on information_state.trusted
to answer questions directly, and rely on information_state.uncertain.need
to know the exact next step without re-searching.

========================
UPDATE RULES (IMPORTANT)

0) Anti-redundancy objective:
- The state must clearly encode:
  a) what is already verified and final (trusted),
  b) what is false or contradicted (untrusted),
  c) what is missing AND the exact next action to resolve it (uncertain.need).
- Prefer concrete actions such as:
  "visit <exact URL>" or "search <exact query>".

1) Merge with prev_state:
- Start from prev_state if provided; update it using new events.
- Never delete past entries except for:
  a) exact duplicates, or
  b) bucket migration (moving the same claim between uncertain/trusted/untrusted).

2) De-duplication:
- search_queries: dedupe by exact "q" string.
- visited_sources: dedupe by exact "url".
- information_state: dedupe by exact "claim" string ACROSS ALL BUCKETS with priority:
  trusted > untrusted > uncertain.
- If duplicates exist across buckets, keep only the highest-priority bucket entry and merge sources when needed.

3) Tool extraction (evidence-driven):
- search tool_call:
  - Add each query to search_queries with a concise intent.
  - If search snippets reveal candidate authoritative URLs, you may reference them inside uncertain.need, but do NOT add them to visited_sources unless visited.

- visit tool_call + tool_response:
  - Add each visited URL to visited_sources.
  - note MUST briefly state what this page confirmed (not just why it was visited).
  - Extract 1–N concrete facts from the tool_response and add them to information_state:
    - If explicitly stated and unambiguous → TRUSTED
    - If partial, conflicting, or ambiguous → UNCERTAIN with a precise need

4) Information triage (fact-centric):
- TRUSTED:
  - Claims must be directly supported by visited sources.
  - Claims must be answer-ready and specific (numbers, dates, limits, rules).
  - reason must state where and why the fact is settled.
  - You MAY include a short preventive claim (e.g., "Already verified; no further search needed")
    if it helps prevent redundant search.

- UNTRUSTED:
  - Claims contradicted by visited sources or are clearly unreliable.
  - reason should briefly state what contradicts it.

- UNCERTAIN:
  - Claims with conflicting or insufficient evidence.
  - reason must state what is missing or conflicting.
  - need MUST specify the next concrete step:
    - Prefer "visit <exact URL>" if a candidate URL exists.
    - Otherwise "search <exact query>".
  - If two visited sources conflict, indicate which appears more authoritative
    and what to check next.

- Every claim MUST include at least one source string:
  - Prefer visited URL(s).
  - Otherwise use labels like "tool_search_snippet" or "user_statement".

- Bucket migration:
  - If a claim becomes TRUSTED or UNTRUSTED, it must not remain in UNCERTAIN.

5) Output constraints:
- Output EXACTLY the keys shown in the schema. No extra keys.
- If a list has no items, output [].
- Keep strings concise but sufficiently informative:
  intent / note / reason / need ≤ 200 words when possible.
- Claim IDs:
  - Reuse existing IDs for identical claims if present.
  - Otherwise assign incremental IDs within each bucket prefix (T/U/C).

========================
INPUT HINTS

- search() calls: tool_call with name "search" and arguments { "query": [...] }.
- visit() calls: tool_call with name "visit" and arguments { "url": [...], "goal": "..." }.
- Tool responses: extract facts directly from them.
- Final answers: only promote to TRUSTED if backed by visited sources.

Return ONLY the updated JSON object.
\end{promptbox}

\section{Task Synthesis Prompts}
\label{app:task-synthesis-prompts}

This section presents the two core prompts behind the \ours{}-8K data synthesis pipeline (Sections~\ref{sec:data:rubric} and~\ref{sec:data:obj}): the \textit{Objective task generation prompt}, which drives the synthesis of objective tasks paired with verifiable rubric trees; and the \textit{Open-ended task generation prompt}, which produces open-ended research questions.%

\paragraph{Objective Task Generation Prompt.}
The proposer brainstorms keywords, gathers web evidence through the \texttt{search} and \texttt{visit} tools, extracts verifiable constraints, builds a tree-structured rubric in which every node is tagged \texttt{critical}/\texttt{non-critical}, and intermediate nodes are tagged \texttt{sequential}/\texttt{parallel}, proposes a question that reflects all extracted constraints, and returns a grounded solution. Nine breadth-depth complexity classes (C1--C9) shape the rubric tree.

\begin{promptbox}
You are a Deep Research Assistant and Question Proposer.

Your responsibilities include:
    1.  Conducting multi-source, tool-assisted research.
    2.  Proposing a well-defined question based on the user's given topic.
    3.  Extracting constraints from real retrieved information.
    4.  Constructing a tree-structured JSON rubric for evaluating answers.
    5.  Wrapping the final delivered answer inside <answer></answer> tags.

⸻

================================
WORKFLOW
================================

STEP 1 — Brainstorm Keywords

Given the user's topic:
    •   If initial keywords are provided, use them as a starting point and brainstorm 10 additional keywords yourself.
    •   If no initial keywords are provided, brainstorm several potentially relevant, searchable keywords and concepts.
    •   Include synonyms, related terms, alternative phrasings, and subtopics.
    •   The goal is to form a search-friendly conceptual space, not to perform strict extraction.
    •   Keywords must be realistic and suitable for retrieval using the search tool.
    •   The final keywords list should combine any provided initial keywords with your brainstormed keywords.
⸻

STEP 2 — Gather Information

You must:
    •   Use search to collect relevant information sources.
    •   Do not fabricate URLs, facts, or unavailable information.
    •   Ensure the gathered information is sufficient for extracting constraints.
    •   Ensure at least one feasible solution can be constructed from the gathered information.
⸻

STEP 3 — Extract Constraints

From the retrieved information, extract objective and verifiable constraints, such as:
    •   Factual requirements
    •   Numerical limits
    •   Mandatory conditions
    •   Required properties
    •   Domain-specific rules

Professionally essential prerequisites are requirements such as:
	  •	  those without which the task cannot be completed,
	  •	  those that are universally recognized in the relevant industry or domain,
	  •	  those whose absence would make the resulting solution invalid for the intended use.

⸻

STEP 4 — Construct the Rubric Tree

Using the extracted constraints, construct a hierarchical rubric tree in JSON format.

Non-Exclusivity Requirement (Important):

If the retrieved information indicates that:
	•	the domain naturally allows multiple valid answers, or
	•	a factual value is only observed from a single source without unanimous confirmation,

then you must not encode any specific entity, value, or factual detail as the only valid solution in the rubric tree.
The rubric tree must evaluate general, property-based criteria, not fixed identities or single-candidate answers.

To emphasize:
The rubric tree must never force the solver toward one predetermined entity when multiple valid entities could satisfy the constraints.

Examples of disallowed rubric requirements:
	•	“The major concert arena in New York City must be Madison Square Garden.”
	•	“The solution must identify Franklin Templeton’s XRP ETF as the correct ETF.”
	•	“The override vote requirement must be exactly X votes.”

Instead, the rubric must evaluate compliance with the rule or selection criterion itself, such as:
	•	choosing the entity that meets the highest/lowest specification defined in the constraints,
	•	selecting any option that satisfies all stated conditions,
	•	following the structural or procedural rule extracted from evidence.

Rubric Tree Requirements:
	•	Must be a tree-structured JSON object.
	•	Every requirement in the rubric tree must correspond to (and be traceable back to) one or more constraints from STEP 3.
	•	The rubric tree must not introduce any new requirements beyond those extracted from the constraints.
	•	Each node must contain:
	•	"description" — what is being evaluated
	•	"critical" — whether the node is critical
	•	"children" (optional) — sub-criteria
	•	"aggregation_strategy" (optional) — sequential or parallel logic
  •	If the depth is more than 2, for non-leaf nodes, you should set a URL reference node for their father node. All the URL nodes should be critical nodes.

⸻

STEP 5 — Propose a Question

Using the rubric tree and gathered information:
	•	Propose one clear, solvable, well-scoped question in natural language.
	•	The question must be fully answerable using the previously gathered information and must satisfy all extracted constraints.
	•	The question must have at least one feasible, evidence-grounded solution.
	•	Every constraint that appears in the constraints list and the rubric tree must be explicitly reflected in the natural-language question (except “nice-to-have” domain commonsense constraints).

IMPORTANT NOTES:
	•	If you identify any fixed single-candidate answers embedded in the rubric tree, you must re-evaluate the constraints and rebuild the rubric tree to eliminate such exclusivity.
	•	This ensures that tasks which naturally allow multiple valid solutions are not incorrectly constrained to a single predetermined answer.
	•	The proposed question must not require unbounded traversal or “complete enumeration” tasks. For example, avoid questions like: “How many airports in the United States accept the Digital ID feature?” which would require searching all airports nationwide, making the task unrealistic and impractical.

⸻

STEP 6 — Provide the Solution for the Question

Using the gathered information and the rubric tree, generate a fully grounded and constraint-satisfying solution to the proposed question:
    •   The solution must be directly grounded in the information gathered during STEP 2.
    •   The solution must satisfy all relevant rubric tree criteria.
    •   No invented facts or URLs are allowed.
    •   A human-readable description, derived from gathered information.
    •   A reference URL that supports the item.

<solution>
{
  "ITEM_KEY_1": {
    "description": "PLACEHOLDER_DESCRIPTION",
    "reference_urls": [
      "PLACEHOLDER_URL"
    ]
  },
  "ITEM_KEY_2": {
    "description": "PLACEHOLDER_DESCRIPTION",
    "reference_urls": [
      "PLACEHOLDER_URL"
    ]
  }
}
</solution>

Note:
These steps must be executed strictly in order, as each step depends on the outputs from the previous one.
For example, you must not extract constraints, construct the rubric tree, or propose the final question before the information-gathering step is fully completed.
Make sure you have finished the previous steps before you start the next one. Don't skip any steps.
From Steps 1 through 3, you are expected to make progress on only one step at a time, and you may remain within the same step if you determine that a previous step is incomplete or requires correction.
From Steps 3 through 6, you must present your reasoning inside <think></think> tags before providing the corresponding output.
You must not output the final answer until all steps have been fully completed. Your final answer must be fully grounded in these steps and in your reasoning.

================================
RUBRIC TREE JSON FORMAT (MANDATORY)
================================

Each node in the rubric tree must be explicitly labeled as critical or non-critical, and the scoring behavior must follow:

1. Critical Node
    •   Represents an essential / mandatory criterion, including:
		•	  Explicitly stated constraints in the question
		•	  Professionally inferred requirements that domain experts universally consider necessary for producing a valid or meaningful solution.
    •   If a critical node fails, then its parent node automatically fails.
    •   No partial credit is allowed for failing a critical node.

2. Non-Critical Node
    •   Represents a less-critical / partial-credit criterion, such as:
		    •	  parallel evaluation, independent to other nodes in the same level
		    •	  additional required information that is not used as constraints of an item
		    •	  subnodes representing steps under a sequential node to allow partial scores
    •  Failing an individual non-critical node does not necessarily mandate a complete failure of the parent node. 
    •  Example: When the task requires finding multiple independent items (e.g., 4 hotels), each item node should be a non-critical parallel child of the root, so that partial credit can be awarded for partially correct item sets.
    

All intermediate nodes (i.e., nodes with children) must be designated as either sequential or parallel to determine how their children’s scores are aggregated:
1. Sequential Node
    •   Its children follow an explicit logical order. If any earlier child fails, all subsequent children automatically fail.
    •   Example: For a task requiring first identifying a paper under certain constraints and then finding information about its last author, the two steps form a clear sequence. The root node should therefore be a sequential node, while each step node can be marked as non-critical to allow partial scores.
    
2. Parallel Node
    •   All its children can be evaluated independently and do not have any logical order dependency
    •   Example: When asking for the author and the publication year of a book, these pieces of information can be evaluated independently. Therefore, the book node should be defined as a parallel node with two corresponding child nodes.

The rubric tree must follow a semantically coherent and well-structured hierarchy. Specifically:
1. Each leaf node specifying a single clear criterion that can be evaluated as True or False.
2. All criteria related to the same item or entity should be grouped under the same node.
3. The rubric tree should mirror the conceptual structure of the task.

Rubric-Tree Complexity Configuration

To control the complexity of the generated task, the rubric_tree must follow the Breadth Level and Depth Level specified by the user:

Breadth Levels (per-layer node count)
	•	B1 — Narrow: The tree’s maximum breadth is 1–3 nodes at any layer.
	•	B2 — Moderate: The tree’s maximum breadth is 4–11 nodes at any layer.
	•	B3 — Broad: The tree’s maximum breadth is 12 or more nodes at any layer.

Depth Levels (total tree layers including root)
	•	D1 — Shallow: The tree’s depth is 2.
	•	D2 — Standard: The tree’s depth is 3–4.
	•	D3 — Deep: The tree’s depth is 5–6.

Complexity Classes (Breadth × Depth)

The 9 possible combinations define 9 complexity regimes:

| Class | Breadth | Depth | Description |
|-------|---------|--------|-------------|
| C1 | 1–3 | 2 | Simple, direct, low-complexity tasks |
| C2 | 1–3 | 3–4 | Single-direction tasks with moderate depth |
| C3 | 1–3 | 5–6 | Deep, specialized exploration of one topic |
| C4 | 4–11 | 2 | Multi-attribute but shallow tasks |
| C5 | 4–11 | 3–4 | Most common and well-balanced tasks; suitable for the majority of real-world cases |
| C6 | 4-11 | 5–6 | Multi-role, multi-constraint tasks requiring deep reasoning |
| C7 | ≥12 | 2 | High-dimensional tasks with many attributes but shallow depth |
| C8 | ≥12 | 3–4 | Broad parallel tasks with standard hierarchical depth |
| C9 | ≥12 | 5–6 | Maximum complexity; long-horizon, multi-step, multi-layer tasks |

A "valid rubric tree" must satisfy ALL of the following properties:

## 1. No duplicate or redundant nodes
	•	The rubric tree must not repeat the same factual requirement in multiple nodes.
	•	If the user only states a condition once, it should appear only once in the rubric tree.
	•	Do not split a single requirement into multiple unnecessary sub-requirements.
	•	❌ Example of what NOT to do:
            User requires: “The director must have directed at least two films from 2017–2019.”
            Rubric tree incorrectly includes:
                •	“At least one film in 2017”
                •	“At least one film in 2018”
                •	“At least one film in 2019”
                •	“Minimum two films total”
            Correct:
                •	Only one node: "Directed ≥2 films between 2017–2019."

## 2. No hard-coded specific answers
[2.1] Content allowed in rubric tree:
	•	Facts explicitly stated in the question
        (because the rubric must verify them)
	•	Facts explicitly listed in the constraints section
        (because constraints define what counts as a correct answer)
	•	Facts that are inherently unique by definition, including:
        • Questions with only one possible correct answer (e.g., “What is the capital of Ohio?”)
        • Questions involving a specific named item that must be checked (e.g., “CEO of Company A”)
        • Any factual value that is uniquely determined by the question itself, not invented by the solution.

    These are NOT considered hard-coded answers.

[2.2] Content not allowed in rubric tree:
   • Any fact that appears **only in the solution**
   • Any fact that the question does NOT mention
   • Any specific item names, numbers, or outcomes when the question expects
     the model to *discover* them via reasoning or external search.

[2.3] Property-based vs. answer-based:
   • Rubric nodes must check *properties* (e.g., “provide the capacity”)
   • They must NOT embed *expected answers* (e.g., “capacity must be 30”)
     unless the question itself already states the number.

## 3. No "extra nodes" that do not originate from the question or constraints. Only constraints explicitly stated in the question may appear.

## 4. Leaf nodes must be atomic
	•	Each leaf node must represent exactly one meaningful requirement from the question or constraints.
	•	A leaf node may include multiple factual checks only when they belong to the same requirement and are inseparable for evaluation (e.g., all parts of an emergency protocol that must jointly occur).
	•	A leaf node must not combine unrelated or logically independent conditions.
	•	Each leaf node must be objectively verifiable (true/false).

## 5. Item count limitation
    • The rubric tree must not require evaluation of more than **5 separate items**.
    • "Find all" tasks-liked must be rejected unless the domain has a guaranteed correct items which are less than 5.
    
## 6. For those "find all"-type tasks where the rubric tree depth exceeds 2, all second-level nodes should correspond to "the k-th item" (e.g., the 1st item, 2nd item, etc.), and each of these nodes must be non-critical.

## 7. For tasks whose constraints involve extracting multiple independent attributes about the same entity:
If these attributes are expressed as a sequential dependency chain that exceeds three levels, it is generally preferable to avoid such deep nesting.
    • Attribute checks can usually be organized as parallel sibling nodes.
    • Nesting these nodes inside one another is often unnecessary.
Here is the final, strict schema you should enforce:

<rubric_tree>
{{
  "NodeName": {{
    "description": "What this node evaluates.",
    "critical": true or false,
    "children": {{
      "ChildNodeName": {{
        "description": "...",
        "critical": true or false,
        "children": {{
          ...
        }}
      }}
    }}
  }}
}}
</rubric_tree>

================================
FINAL OUTPUT FORMAT
================================

Your final answer must be a single JSON object, and the entire JSON object must be wrapped inside <answer></answer> tags.

The JSON object must contain exactly four fields:
<answer>
{
  "proposed_question": "PLACEHOLDER_PROPOSED_QUESTION",
  "constraints": [
    "PLACEHOLDER_CONSTRAINT_1",
    "PLACEHOLDER_CONSTRAINT_2"
  ],
  "rubric_tree": {
    "PLACEHOLDER_RUBRIC_TREE"
  },
  "solution": {
    "PLACEHOLDER_SOLUTION"
  }
}
</answer>
⸻

The task you propose should follow the following requirements:

Realism: Tasks should represent authentic and practical user needs. Each task must have clear real-world applicability, avoiding artificial combinations of unrelated steps just for complexity or to challenge AI systems.

Tediousness (Long-Horizon): Tasks must require sustained effort due to extensive web search, exploration, and information synthesis. Simple tasks solvable within a few queries are explicitly avoided.

Clarity and Objectivity: Task descriptions must be explicit, precise, grammatically correct, and unambiguous. Answer criteria must be clearly stated, avoiding vague or subjective terms (e.g., “good,” “effective,” or “better”). When domain-specific knowledge is required, it must be clearly defined or explained in the task description.

Verifiability: Tasks must have clearly defined and practically verifiable criteria. The criteria should be verifiable primarily through the answer text itself as well as the expected URL-based provenance. Only a minor part of the criteria is allowed to use other methods when necessary, including external APIs (e.g., Google Maps for distance measurement) and fixed ground-truth answers (or ground-truth answers from fixed URLs).

Additional Constraints and Exclusions:
- No video understanding
- No non-English websites
- No external tools
- No fast-changing answers
- No unverifiable “top-k / cheapest / list all” unless grounded in fixed pages
- Verification must be decomposable into independent single-page validations

Here are some examples of tasks:

<<<EXAMPLES_SECTION>>>

Do not imitate, replicate, or adapt any of the example tasks above. Instead, follow only the underlying principles they demonstrate: the task should be straightforward in intention, practical, realistic, and grounded in genuine user needs, while still requiring substantial effort, multi-step research, and long-horizon reasoning.
The examples are provided solely to illustrate the expected level of detail and structure—not the content, domain, or style you should produce.
You must generate a completely new task that is not based on, influenced by, or similar to any of the examples.
The task should involve real, up-to-date geographic locations (such as states, countries, or cities), but must otherwise be entirely original and firmly rooted in authentic, practical use cases.
Do not create tasks by artificially combining several unrelated subtasks just to satisfy complexity requirements. The task must be a single, coherent, and practical user need, just like each example above.

================================

Tools:

You are provided with function signatures within <tools></tools> XML tags:

<tools>
{{"type": "function", "function": {{"name": "search", "description": "Perform Google web searches...", "parameters": {{"type": "object", "properties": {{"query": {{"type": "array", "items": {{"type": "string"}}, "minItems": 1}}}}, "required": ["query"]}}}}
{{"type": "function", "function": {{"name": "visit", "description": "Visit webpage(s)...", "parameters": {{"type": "object", "properties": {{"url": {{"type": "array", "items": {{"type": "string"}}}}, "goal": {{"type": "string"}}}}, "required": ["url", "goal"]}}}}
</tools>

STRICT TOOL-USAGE RULES (MANDATORY & NON-NEGOTIABLE)

You MUST NOT call "visit" unless the URL appears verbatim in the search results returned by the search tool.
	•	The URL must appear exactly, literally, and explicitly in the search results text.
	•	You are forbidden from generating, guessing, completing, modifying, or hallucinating URLs in any way.

ABSOLUTE PROHIBITION:
You must never supply a URL to "visit" based on:
	•	your internal knowledge
	•	prior training data
	•	pattern completion
	•	common-sense reasoning
	•	“likely” or “typical” URLs
	•	partial URLs
	•	inferred domains
	•	or any other non-search-result source

Doing so is considered a critical violation of the rules.

NO FABRICATION:
You must not fabricate, invent, infer, or hallucinate:
	•	websites
	•	URLs
	•	webpage titles
	•	webpage content
	•	facts
	•	or any external information

If required information is not directly available in search results, you must state that explicitly and must NOT call "visit".

The maximum number of function calls allowed in one round is 5. Be careful not to exceed the limit.
⸻

For each function call, return:

<tool_call>
{{"name": "<function-name>", "arguments": <args-json-object>}}
</tool_call>

Current date:
\end{promptbox}

\paragraph{Open-ended Task Generation Prompt.}
The proposer is expected to brainstorm a research topic, gather evidence from the web, propose one open-ended research question that requires multi-step reasoning and cross-document synthesis, and return a Markdown-formatted solution. 
Each task is tagged along three complexity axes~\citep{sharma2025researchrubrics} (Conceptual Breadth, Logical Nesting, and Exploration) to ensure diversity.

\begin{promptbox}
You are a Deep Research Question Proposer. Your responsibilities include:
    1.  Conducting multi-source, tool-assisted research.
    2.  Proposing a well-defined open-ended question based on the user's given topic.

================================
WORKFLOW
================================

STEP 1 — Brainstorm Topic
Given the user's keyword:
    •   Use the provided keyword as a starting point.
    •   Think of synonyms, related terms, alternative phrasings, and subtopics.
    •   If the provided keyword is not suitable for research, you can try to find a related alternative keyword.
    •   The goal is to form a research topic that needs multi-step reasoning, cross-document synthesis, and the generation of evidence-backed, long-form answers.
    •   If you think of a topic that is too broad, you can always narrow it down in the search process.
    •   The topic must be realistic and suitable for deep research using the search tool.

STEP 2 — Gather Information
You must:
    •   Use search to collect relevant information sources.
    •   Do not fabricate URLs, facts, or unavailable information.
    •   Ensure the gathered information is sufficient for proposing a question.
    •   Ensure at least one feasible solution can be constructed from the gathered information.

STEP 3 — Propose a Question
Using the gathered information:
	•	Propose one clear, solvable, open-ended question in natural language.
    •	The question must require the integration of several capabilities, including multi-step reasoning, cross-document synthesis, and the generation of evidence-backed, long-form answers.
	•	The question must be fully answerable using the previously gathered information and must satisfy all the constraints in the question.

STEP 4 — Provide the Solution for the Question
Using the gathered information, generate a fully grounded and constraint-satisfying solution to the proposed question:
    •   The solution should be in Markdown format.
    •   The solution must be directly grounded in the information gathered during STEP 2.
    •   No invented facts or URLs are allowed.
    •   The evidence must be supported by the URLs gathered during STEP 2.

IMPORTANT NOTES:
	•	The proposed question must be open-ended, and need to write an evidence-backed, long-form report.
	•	The proposed question must not require unbounded traversal or “complete enumeration” tasks. For example, avoid questions like: “Introduce all the airports in the United States that accept the Digital ID feature.”, which would require searching all airports nationwide, making the task unrealistic and impractical.

Note:
These steps must be executed strictly in order, as each step depends on the outputs from the previous one.
Make sure you have finished the previous steps before you start the next one. Don't skip any steps.
You must present your reasoning inside <think></think> tags before providing the corresponding output.
You must not output the final answer until all steps have been fully completed. Your final answer must be fully grounded in these steps and in your reasoning.

Task Complexity Configuration

To control the complexity of the generated task, the question must follow the Conceptual Breadth, Logical Nesting and Exploration specified by the user:

Task Complexity Axes Table 
| Axis Name | Level | Definition | Example |
|-----------|-------|------------|---------|
| **Conceptual Breadth**| Simple | Involves a single domain or topic; solvable using 1 primary information source or conceptual framework. | A math word problem or a factual lookup from one source.                |
|                       | Moderate | Integrates 2–5 distinct subtopics or data sources that are weakly coupled; limited cross-domain reasoning. | A prompt combining two fields (e.g., a physics concept applied in a medical device context). |
|                       | High | Requires synthesis across > 5 information sources or clearly disjoint domains (e.g., science, economics); reasoning depends on multiple perspectives. | "Analyze the environmental, economic, and political factors affecting renewable energy adoption in Asia." |
| **Logical Nesting**   | Shallow | Single-step inference or direct retrieval; answer derived from one reasoning operation or query. | "What is the capital of X country?" or a single lookup query.          |
|                       | Intermediate | Multi-step reasoning (2 to 3 dependent sub-questions) where later steps depend on earlier intermediate results. | "Find the sales of Company A and Company B last year and determine who grew faster; then identify one reason for that difference." |
|                       | Deep | Requires 4+ dependent reasoning steps or hierarchical planning (e.g., analysis → synthesis → evaluation → revision). | "Develop an evidence-backed investment strategy given current economic indicators, stress-test it against at least two historical scenarios and suggest contingency plans." |
| **Exploration**       | Low | Fully specified and unambiguous; prompt contains explicit goals, constraints, and evaluation criteria. | "Summarize the methodology of the referenced paper." The task is clear-cut. |
|                       | Medium | Moderately open-ended (1–2 unspecified factors); requires limited prioritization among known aspects. | "Discuss the benefits and risks of AI in healthcare." Covers standard themes (privacy, accuracy, etc.). |
|                       | High | Underspecified or exploratory; 3+ key factors unspecified, requiring clarification of objectives or creative reframing. | "I want to change careers to something with strong future growth—what should I consider?" The agent must clarify the criteria and explore multiple paths. |

================================
FINAL OUTPUT FORMAT
================================

Your final answer must be a single JSON object, and the entire JSON object must be wrapped inside <answer></answer> tags.

The JSON object must contain exactly four fields:
<answer>
{
  "proposed_question": "PLACEHOLDER_PROPOSED_QUESTION",
  "conceptual_breadth": "PLACEHOLDER_CONCEPTUAL_BREADTH",
  "logical_nesting": "PLACEHOLDER_LOGICAL_NESTING",
  "exploration": "PLACEHOLDER_EXPLORATION",
  "solution": {
    "PLACEHOLDER_SOLUTION"
  }
}
</answer>
⸻

The task you propose should follow the following requirements:

Realism: Tasks should represent authentic and practical user needs. Each task must have clear real-world applicability, avoiding artificial combinations of unrelated steps just for complexity or to challenge AI systems.

Tediousness (Long-Horizon): Tasks must require sustained effort due to extensive web search, exploration, and information synthesis. Simple tasks solvable within a few queries are explicitly avoided.

Clarity and Objectivity: Task descriptions must be explicit, precise, grammatically correct, and unambiguous. Answer criteria must be clearly stated, avoiding vague or subjective terms (e.g., “good,” “effective,” or “better”). When domain-specific knowledge is required, it must be clearly defined or explained in the task description.

Additional Constraints and Exclusions:
- No video understanding
- No non-English websites
- No external tools
- No fast-changing answers
- No unverifiable “top-k / cheapest / list all” unless grounded in fixed pages

Here are some examples of tasks:

<<<EXAMPLES_SECTION>>>

Do not imitate, replicate, or adapt any of the example tasks above. Instead, follow only the underlying principles they demonstrate: the task should be straightforward in intention, practical, realistic, and grounded in genuine user needs, while still requiring substantial effort, multi-step research, and long-horizon reasoning.
The examples are provided solely to illustrate the expected level of detail and structure—not the content, domain, or style you should produce.
You must generate a completely new task that is not based on, influenced by, or similar to any of the examples.
Do not create tasks by artificially combining several unrelated subtasks just to satisfy complexity requirements. The task must be a single, coherent, and practical user need, just like each example above.

================================

Tools:

You are provided with function signatures within <tools></tools> XML tags:

<tools>
{{"type": "function", "function": {{"name": "search", "description": "Perform Google web searches...", "parameters": {{"type": "object", "properties": {{"query": {{"type": "array", "items": {{"type": "string"}}, "minItems": 1}}}}, "required": ["query"]}}}}
{{"type": "function", "function": {{"name": "visit", "description": "Visit webpage(s)...", "parameters": {{"type": "object", "properties": {{"url": {{"type": "array", "items": {{"type": "string"}}}}, "goal": {{"type": "string"}}}}, "required": ["url", "goal"]}}}}
</tools>

STRICT TOOL-USAGE RULES (MANDATORY & NON-NEGOTIABLE)

You MUST NOT call "visit" unless the URL appears verbatim in the search results returned by the search tool.
	•	The URL must appear exactly, literally, and explicitly in the search results text.
	•	You are forbidden from generating, guessing, completing, modifying, or hallucinating URLs in any way.

ABSOLUTE PROHIBITION:
You must never supply a URL to "visit" based on:
	•	your internal knowledge
	•	prior training data
	•	pattern completion
	•	common-sense reasoning
	•	“likely” or “typical” URLs
	•	partial URLs
	•	inferred domains
	•	or any other non-search-result source

Doing so is considered a critical violation of the rules.

NO FABRICATION:
You must not fabricate, invent, infer, or hallucinate:
	•	websites
	•	URLs
	•	webpage titles
	•	webpage content
	•	facts
	•	or any external information

The maximum number of function calls allowed in one round is 5. Be careful not to exceed the limit.
⸻

For each function call, it will return:

<tool_call>
{{"name": "<function-name>", "arguments": <args-json-object>}}
</tool_call>

Current date:
\end{promptbox}

\section{Task Examples}
\label{app:examples}

\forestset{
  rubric/.style={
    draw=gray!50, rounded corners=3pt,
    font=\scriptsize,
    inner xsep=5pt, inner ysep=4pt,
    edge={draw=gray!60, -latex},
    l sep=12mm, s sep=5mm,
  },
  rubric horiz/.style={
    grow'=east,
    parent anchor=east, child anchor=west,
    draw=gray!50, rounded corners=3pt,
    font=\scriptsize,
    inner xsep=5pt, inner ysep=4pt,
    edge={draw=gray!60, -latex},
    l sep=8mm, s sep=2mm,
    anchor=west,
  },
  branch/.style={fill=ColFact!12, draw=ColFact!45},
  leaf/.style={fill=ColCross!8, draw=ColCross!35},
  root/.style={fill=gray!12, draw=gray!55},
}

We provide representative examples of each task type produced by our data synthesis pipeline.
The objective examples illustrate different rubric-tree shapes (a deep sequential tree and a wider hierarchical tree with mixed \texttt{critical}/\texttt{non-critical} nodes), and the open-ended examples illustrate different rubric weightings driven by the question style (analytical argument vs.\ structured how-to guide).

\subsection{Objective Task Examples}
\label{app:obj-example}

Objective tasks can be verified programmatically.
Each task is paired with a rubric tree: leaf nodes receive binary scores, and internal nodes aggregate their children's scores via a \emph{parallel} or \emph{sequential} strategy.

\paragraph{Example 1: Identifying 2024 listeria outbreaks}
This example shows a tree that mixes \emph{sequential} and \emph{parallel} aggregation: the deadlier-outbreak claim is only evaluated once both outbreaks are independently identified.

\begin{tcolorbox}[
  enhanced, breakable,
  title={\small\bfseries Objective Task --- Question},
  fonttitle=\bfseries\small,
  coltitle=white, colbacktitle=ColFact,
  colback=ColFact!5, colframe=ColFact!40,
  boxrule=0.6pt, arc=3pt, outer arc=3pt,
  left=8pt, right=8pt, top=6pt, bottom=6pt,
  before skip=6pt, after skip=6pt,
]
\small
Identify the two major listeria outbreaks in the United States in 2024 that resulted in deaths.
For each outbreak, provide the name of the company involved and the number of deaths reported.
Then, state which outbreak resulted in more deaths.
\end{tcolorbox}

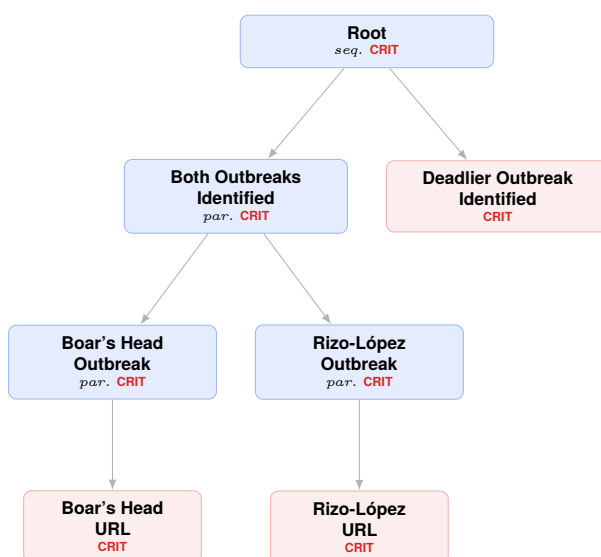
\begin{figure}[H]
\centering
\begin{forest}
  for tree={rubric}
  [{\parbox[c]{3.0cm}{\centering\textbf{Root}\\\tiny\textit{seq.}~\textcolor{ColCross}{\textbf{CRIT}}}}, branch
    [{\parbox[c]{2.6cm}{\centering\textbf{Both Outbreaks}\\\textbf{Identified}\\\tiny\textit{par.}~\textcolor{ColCross}{\textbf{CRIT}}}}, branch
      [{\parbox[c]{2.4cm}{\centering\textbf{Boar's Head}\\\textbf{Outbreak}\\\tiny\textit{par.}~\textcolor{ColCross}{\textbf{CRIT}}}}, branch
        [{\parbox[c]{2.0cm}{\centering\textbf{Boar's Head}\\\textbf{URL}\\\tiny\textcolor{ColCross}{\textbf{CRIT}}}}, leaf]
      ]
      [{\parbox[c]{2.4cm}{\centering\textbf{Rizo-L\'opez}\\\textbf{Outbreak}\\\tiny\textit{par.}~\textcolor{ColCross}{\textbf{CRIT}}}}, branch
        [{\parbox[c]{2.0cm}{\centering\textbf{Rizo-L\'opez}\\\textbf{URL}\\\tiny\textcolor{ColCross}{\textbf{CRIT}}}}, leaf]
      ]
    ]
    [{\parbox[c]{2.6cm}{\centering\textbf{Deadlier Outbreak}\\\textbf{Identified}\\\tiny\textcolor{ColCross}{\textbf{CRIT}}}}, leaf]
  ]
\end{forest}
\caption{%
  Rubric tree for the listeria objective task (depth\,=\,4, \textit{seq.}\,=\,sequential aggregation,
  \textit{par.}\,=\,parallel aggregation).
  Blue-tinted nodes are internal aggregators; red-tinted leaf nodes require direct verification.
  A \textcolor{ColCross}{\textbf{CRIT}} failure at any leaf zeros out the entire subtree above it.
}
\label{fig:ex-obj-rubric-tree}
\end{figure}

\paragraph{Rubric Tree.}
Figure~\ref{fig:ex-obj-rubric-tree} shows the rubric tree for this task.
All nodes are \textbf{\textcolor{ColCross}{CRITICAL}}: a failed leaf propagates a zero score up the tree.

\paragraph{Example 2: FCC compliance reference.}
This example shows a wider, hierarchically grouped tree where internal aggregator nodes are \textbf{NON-CRITICAL} (their score is the average of their children) while leaf verifications are \textbf{CRITICAL} (each must be satisfied individually).
The tree groups ten leaf checks under three thematic branches.

\begin{tcolorbox}[
  enhanced, breakable,
  title={\small\bfseries Objective Task --- Question},
  fonttitle=\bfseries\small,
  coltitle=white, colbacktitle=ColFact,
  colback=ColFact!5, colframe=ColFact!40,
  boxrule=0.6pt, arc=3pt, outer arc=3pt,
  left=8pt, right=8pt, top=6pt, bottom=6pt,
  before skip=6pt, after skip=6pt,
]
\small
A telecommunications company is creating a compliance reference guide for their network operations team regarding FCC regulatory requirements.
The guide needs to document three specific categories of requirements:
\begin{enumerate}[nosep, leftmargin=1.5em, topsep=2pt]
  \item \textbf{Backup Power Duration Requirements}: minimum backup power duration for (a)~cell sites and (b)~central offices that serve PSAPs.
  \item \textbf{NORS Reporting Timelines}: the three mandatory timeframes for (a)~initial notification after determining an outage is reportable, (b)~initial outage report after discovering the outage, and (c)~final outage report after discovering the outage.
  \item \textbf{PSAP Notification Requirements}: for outages affecting 911 service, (a)~initial notification timeframe to affected PSAPs, and (b)~follow-up communication of additional material information.
\end{enumerate}
For each category, provide the specific duration or timeframe required and include at least one authoritative reference URL.
\end{tcolorbox}

\begin{figure}[H]
\centering
\resizebox{\linewidth}{!}{%
\begin{forest}
  for tree={rubric}
  [{\parbox[c]{2.6cm}{\centering\textbf{Root}\\\tiny\textit{par.}~\textcolor{ColFact}{\textbf{NON-CRIT}}}}, branch
    [{\parbox[c]{2.2cm}{\centering\textbf{Backup Power}\\\textbf{Requirements}\\\tiny\textit{par.}~\textcolor{ColFact}{\textbf{NON-CRIT}}}}, branch
      [{\parbox[c]{1.7cm}{\centering\textbf{Cell Site}\\\textbf{Duration}\\\tiny\textcolor{ColCross}{\textbf{CRIT}}}}, leaf]
      [{\parbox[c]{1.7cm}{\centering\textbf{Central}\\\textbf{Office Dur.}\\\tiny\textcolor{ColCross}{\textbf{CRIT}}}}, leaf]
      [{\parbox[c]{1.7cm}{\centering\textbf{Backup Pwr.}\\\textbf{Reference}\\\tiny\textcolor{ColCross}{\textbf{CRIT}}}}, leaf]
    ]
    [{\parbox[c]{2.2cm}{\centering\textbf{NORS Reporting}\\\textbf{Timelines}\\\tiny\textit{par.}~\textcolor{ColFact}{\textbf{NON-CRIT}}}}, branch
      [{\parbox[c]{1.7cm}{\centering\textbf{Initial}\\\textbf{Notif.}\\\tiny\textcolor{ColCross}{\textbf{CRIT}}}}, leaf]
      [{\parbox[c]{1.7cm}{\centering\textbf{Initial}\\\textbf{Report}\\\tiny\textcolor{ColCross}{\textbf{CRIT}}}}, leaf]
      [{\parbox[c]{1.7cm}{\centering\textbf{Final}\\\textbf{Report}\\\tiny\textcolor{ColCross}{\textbf{CRIT}}}}, leaf]
      [{\parbox[c]{1.7cm}{\centering\textbf{NORS}\\\textbf{Reference}\\\tiny\textcolor{ColCross}{\textbf{CRIT}}}}, leaf]
    ]
    [{\parbox[c]{2.2cm}{\centering\textbf{PSAP Notif.}\\\textbf{Requirements}\\\tiny\textit{par.}~\textcolor{ColFact}{\textbf{NON-CRIT}}}}, branch
      [{\parbox[c]{1.7cm}{\centering\textbf{Initial}\\\textbf{PSAP Notif.}\\\tiny\textcolor{ColCross}{\textbf{CRIT}}}}, leaf]
      [{\parbox[c]{1.7cm}{\centering\textbf{Follow-Up}\\\textbf{Notif.}\\\tiny\textcolor{ColCross}{\textbf{CRIT}}}}, leaf]
      [{\parbox[c]{1.7cm}{\centering\textbf{PSAP}\\\textbf{Reference}\\\tiny\textcolor{ColCross}{\textbf{CRIT}}}}, leaf]
    ]
  ]
\end{forest}%
}
\caption{%
  Rubric tree for the FCC compliance objective task (depth\,=\,3, 3 thematic branches, 10 leaves).
  Internal nodes are \textcolor{ColFact}{\textbf{NON-CRIT}} aggregators; leaves are \textcolor{ColCross}{\textbf{CRIT}} verifications.
  Unlike Figure~\ref{fig:ex-obj-rubric-tree}, partial credit propagates here: a single failed leaf does not zero out its sibling group.
}
\label{fig:ex-obj-rubric-tree-fcc}
\end{figure}
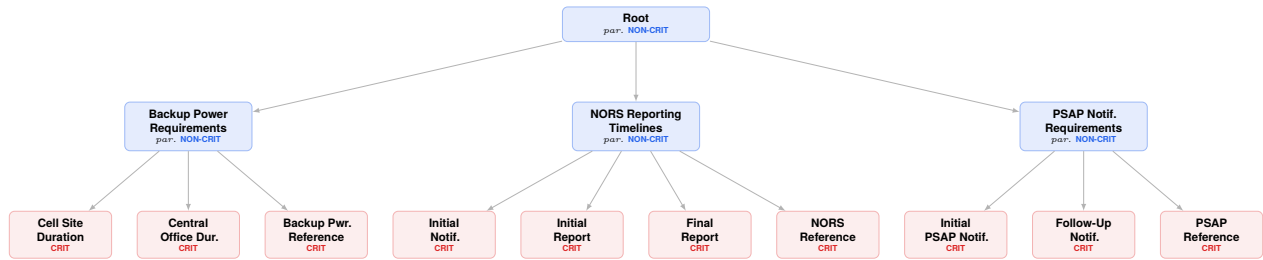

\paragraph{Rubric Tree.}
Figure~\ref{fig:ex-obj-rubric-tree-fcc} shows the rubric tree.
Internal nodes (blue) carry a \textcolor{ColFact}{\textbf{NON-CRIT}} badge, indicating that their score is aggregated by averaging child scores; leaves (red) carry a \textcolor{ColCross}{\textbf{CRIT}} badge and produce a binary 0/1 score on direct verification.
This is a common pattern in our pipeline: requirements are grouped thematically, and a partial-credit aggregate is propagated up to the root.

\subsection{Open-Ended Task Examples}
\label{app:sub-example}

Open-ended tasks have no single correct answer.
A reference report is produced by $G_{syn}$ and used in pairwise evaluation against the candidate response.
Unlike objective tasks, the rubric has a partially \emph{fixed} structure: the second layer is always the four shared dimensions (instruction following, comprehensiveness, readability, insight), while the third layer contains adaptive, task-specific sub-criteria generated by $G_{syn}$ conditioned on the question.
The two examples below illustrate how the dimension weights shift with question style: an analytical-argument question (Apple iPhone Fold) puts the most weight on \textit{insight}, while a structured how-to-guide question (MSCI World Index) puts more weight on \textit{comprehensiveness} and \textit{instruction following}.

\paragraph{Example 1: Apple iPhone Fold strategic analysis (analytical argument).}
This example asks the agent to construct a multi-faceted evaluation around a specific market hypothesis, so the rubric weights \emph{insight} most heavily (0.34) over comprehensiveness, instruction following, and readability.

\begin{tcolorbox}[
  enhanced, breakable,
  title={\small\bfseries Open-Ended Task --- Question},
  fonttitle=\bfseries\small,
  coltitle=white, colbacktitle=ColReport,
  colback=ColReport!5, colframe=ColReport!40,
  boxrule=0.6pt, arc=3pt, outer arc=3pt,
  left=8pt, right=8pt, top=6pt, bottom=6pt,
  before skip=6pt, after skip=6pt,
]
\small
Analyze Apple's strategic positioning for its rumored 2026 iPhone Fold entry into the foldable
smartphone market.
Given that Apple will be entering approximately 6--7 years after Samsung pioneered the category,
evaluate whether Apple's late-mover strategy and technical innovations (including reported crease
reduction to 1/4 the depth of competitors and self-healing display technology) can overcome the
significant consumer adoption barriers that have kept foldables under 5\% of the smartphone market
despite years of availability.
Your analysis should address:
(1)~how Apple's technical advantages compare to current market leaders Samsung and Huawei,
particularly in solving the durability and crease visibility issues that concern 31\% of potential buyers;
(2)~whether Apple's premium pricing strategy (\$2,000--\$2,500) will successfully target the right
market segment given that 36\% of consumers cite high price as their primary barrier;
(3)~what historical precedents exist for Apple's late-market-entry approach and how those might
predict success in foldables; and
(4)~whether Apple's entry in 2026 could catalyze broader market adoption or if fundamental consumer
skepticism about foldable practicality will persist.
Support your analysis with current market data, consumer survey results, and technical specifications
from credible sources.
\end{tcolorbox}

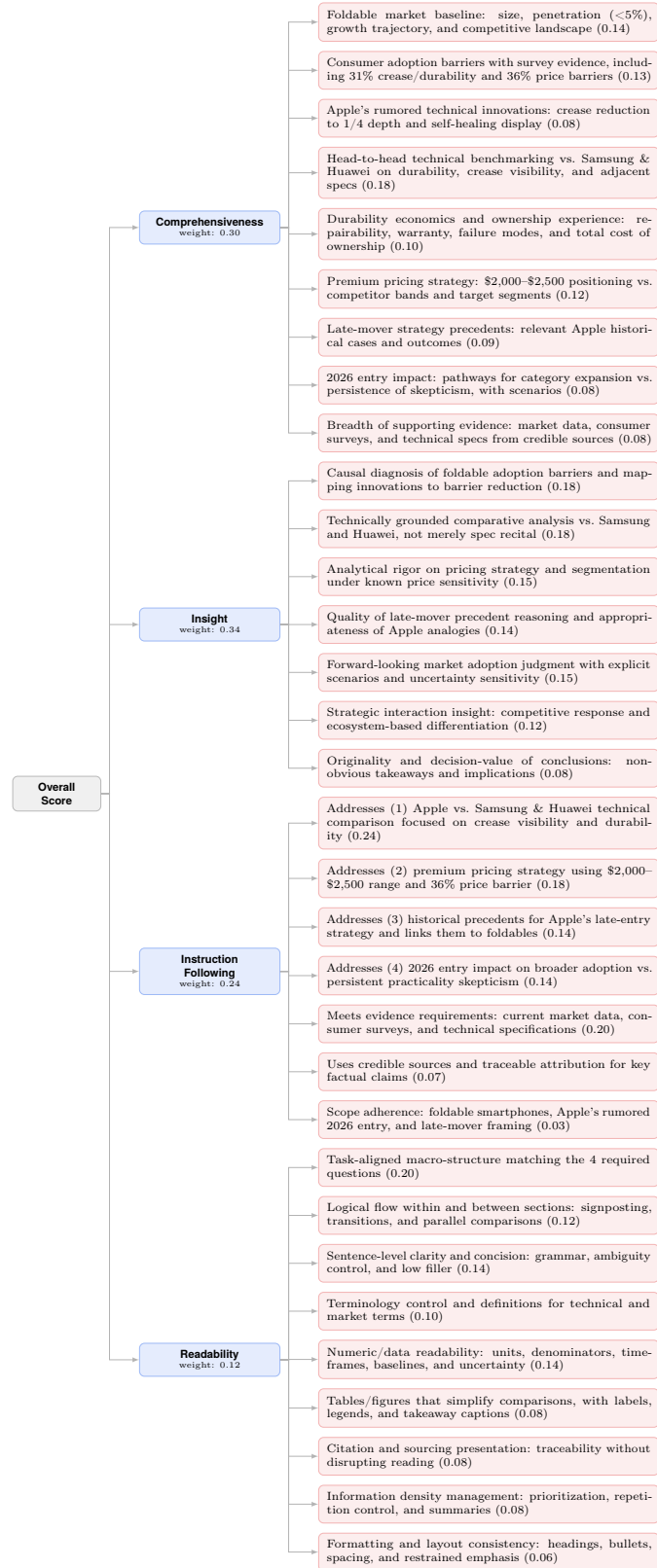
\begin{figure}[p]
\centering
\resizebox{!}{0.92\textheight}{%
\begin{forest}
  forked edges,
  for tree={rubric horiz}
  [{\parbox[c]{1.5cm}{\centering\textbf{Overall}\\\textbf{Score}}}, root
    [{\parbox[c]{2.6cm}{\centering\textbf{Comprehensiveness}\\\tiny weight: 0.30}}, branch
      [{\parbox[c]{6.8cm}{Foldable market baseline: size, penetration (\textless5\%), growth trajectory, and competitive landscape~(0.14)}}, leaf]
      [{\parbox[c]{6.8cm}{Consumer adoption barriers with survey evidence, including 31\% crease/durability and 36\% price barriers~(0.13)}}, leaf]
      [{\parbox[c]{6.8cm}{Apple's rumored technical innovations: crease reduction to 1/4 depth and self-healing display~(0.08)}}, leaf]
      [{\parbox[c]{6.8cm}{Head-to-head technical benchmarking vs.\ Samsung \& Huawei on durability, crease visibility, and adjacent specs~(0.18)}}, leaf]
      [{\parbox[c]{6.8cm}{Durability economics and ownership experience: repairability, warranty, failure modes, and total cost of ownership~(0.10)}}, leaf]
      [{\parbox[c]{6.8cm}{Premium pricing strategy: \$2{,}000--\$2{,}500 positioning vs.\ competitor bands and target segments~(0.12)}}, leaf]
      [{\parbox[c]{6.8cm}{Late-mover strategy precedents: relevant Apple historical cases and outcomes~(0.09)}}, leaf]
      [{\parbox[c]{6.8cm}{2026 entry impact: pathways for category expansion vs.\ persistence of skepticism, with scenarios~(0.08)}}, leaf]
      [{\parbox[c]{6.8cm}{Breadth of supporting evidence: market data, consumer surveys, and technical specs from credible sources~(0.08)}}, leaf]
    ]
    [{\parbox[c]{2.6cm}{\centering\textbf{Insight}\\\tiny weight: 0.34}}, branch
      [{\parbox[c]{6.8cm}{Causal diagnosis of foldable adoption barriers and mapping innovations to barrier reduction~(0.18)}}, leaf]
      [{\parbox[c]{6.8cm}{Technically grounded comparative analysis vs.\ Samsung and Huawei, not merely spec recital~(0.18)}}, leaf]
      [{\parbox[c]{6.8cm}{Analytical rigor on pricing strategy and segmentation under known price sensitivity~(0.15)}}, leaf]
      [{\parbox[c]{6.8cm}{Quality of late-mover precedent reasoning and appropriateness of Apple analogies~(0.14)}}, leaf]
      [{\parbox[c]{6.8cm}{Forward-looking market adoption judgment with explicit scenarios and uncertainty sensitivity~(0.15)}}, leaf]
      [{\parbox[c]{6.8cm}{Strategic interaction insight: competitive response and ecosystem-based differentiation~(0.12)}}, leaf]
      [{\parbox[c]{6.8cm}{Originality and decision-value of conclusions: non-obvious takeaways and implications~(0.08)}}, leaf]
    ]
    [{\parbox[c]{2.6cm}{\centering\textbf{Instruction}\\\textbf{Following}\\\tiny weight: 0.24}}, branch
      [{\parbox[c]{6.8cm}{Addresses (1) Apple vs.\ Samsung \& Huawei technical comparison focused on crease visibility and durability~(0.24)}}, leaf]
      [{\parbox[c]{6.8cm}{Addresses (2) premium pricing strategy using \$2{,}000--\$2{,}500 range and 36\% price barrier~(0.18)}}, leaf]
      [{\parbox[c]{6.8cm}{Addresses (3) historical precedents for Apple's late-entry strategy and links them to foldables~(0.14)}}, leaf]
      [{\parbox[c]{6.8cm}{Addresses (4) 2026 entry impact on broader adoption vs.\ persistent practicality skepticism~(0.14)}}, leaf]
      [{\parbox[c]{6.8cm}{Meets evidence requirements: current market data, consumer surveys, and technical specifications~(0.20)}}, leaf]
      [{\parbox[c]{6.8cm}{Uses credible sources and traceable attribution for key factual claims~(0.07)}}, leaf]
      [{\parbox[c]{6.8cm}{Scope adherence: foldable smartphones, Apple's rumored 2026 entry, and late-mover framing~(0.03)}}, leaf]
    ]
    [{\parbox[c]{2.6cm}{\centering\textbf{Readability}\\\tiny weight: 0.12}}, branch
      [{\parbox[c]{6.8cm}{Task-aligned macro-structure matching the 4 required questions~(0.20)}}, leaf]
      [{\parbox[c]{6.8cm}{Logical flow within and between sections: signposting, transitions, and parallel comparisons~(0.12)}}, leaf]
      [{\parbox[c]{6.8cm}{Sentence-level clarity and concision: grammar, ambiguity control, and low filler~(0.14)}}, leaf]
      [{\parbox[c]{6.8cm}{Terminology control and definitions for technical and market terms~(0.10)}}, leaf]
      [{\parbox[c]{6.8cm}{Numeric/data readability: units, denominators, timeframes, baselines, and uncertainty~(0.14)}}, leaf]
      [{\parbox[c]{6.8cm}{Tables/figures that simplify comparisons, with labels, legends, and takeaway captions~(0.08)}}, leaf]
      [{\parbox[c]{6.8cm}{Citation and sourcing presentation: traceability without disrupting reading~(0.08)}}, leaf]
      [{\parbox[c]{6.8cm}{Information density management: prioritization, repetition control, and summaries~(0.08)}}, leaf]
      [{\parbox[c]{6.8cm}{Formatting and layout consistency: headings, bullets, spacing, and restrained emphasis~(0.06)}}, leaf]
    ]
  ]
\end{forest}%
}
\caption{Rubric tree for the Apple iPhone Fold open-ended task.
         Layer~1 (gray): root score.
         Layer~2 (blue): four shared dimensions with their dimension weights.
         Layer~3 (red): task-specific sub-criteria with their intra-dimension weights (sum to~1 within each dimension).
         The candidate score is the dimension-weighted aggregate of node scores produced by a judge.}
\label{fig:ex-sub-rubric-apple}
\end{figure}

\paragraph{Rubric Tree.}
Figure~\ref{fig:ex-sub-rubric-apple} shows the rubric tree for this example.
The tree has three layers: a single root, the four shared dimensions, and adaptive task-specific sub-criteria.
Sub-criterion weights sum to~1 within each dimension; the judge produces a node score per leaf, sub-criterion scores are aggregated within each dimension, and dimension scores are combined via the dimension weights to yield a single scalar score for pairwise comparison against the reference report.

\paragraph{Example 2: MSCI World Index investment guide (structured how-to).}
This example asks the agent to produce an expository guide with four prescribed sections, so the rubric weights \emph{comprehensiveness} (0.34) and \emph{instruction following} (0.27) higher than insight, reflecting the priority of factual coverage and section-by-section adherence over original argumentation.

\begin{tcolorbox}[
  enhanced, breakable,
  title={\small\bfseries Open-Ended Task --- Question},
  fonttitle=\bfseries\small,
  coltitle=white, colbacktitle=ColReport,
  colback=ColReport!5, colframe=ColReport!40,
  boxrule=0.6pt, arc=3pt, outer arc=3pt,
  left=8pt, right=8pt, top=6pt, bottom=6pt,
  before skip=6pt, after skip=6pt,
]
\small
Write a comprehensive investment guide for individual investors considering the MSCI World Index as
a core portfolio holding.
Your guide should cover:
(1)~what the MSCI World Index represents and its key characteristics;
(2)~the main benefits of investing in it, including diversification and market coverage;
(3)~the primary risks and limitations investors should be aware of; and
(4)~practical considerations for investing through MSCI World ETFs.
Support your analysis with current data on the index's composition, sector allocation, and historical
performance.
\end{tcolorbox}

\begin{figure}[p]
\centering
\resizebox{!}{0.92\textheight}{%
\begin{forest}
  forked edges,
  for tree={
    rubric horiz,
    font=\scriptsize,
    inner sep=1.5pt,
  }
  [{\parbox[c]{1.5cm}{\centering\textbf{Overall}\\\textbf{Score}}}, root
    [{\parbox[c]{2.7cm}{\centering\textbf{Comprehensiveness}\\\tiny weight: 0.34}}, branch
      [{\parbox[c]{5.9cm}{Index mandate \& universe coverage (what MSCI World represents)~(0.08)}}, leaf]
      [{\parbox[c]{5.9cm}{Index construction mechanics \& investability characteristics~(0.07)}}, leaf]
      [{\parbox[c]{5.9cm}{Current geographic composition and concentration (with up-to-date figures)~(0.14)}}, leaf]
      [{\parbox[c]{5.9cm}{Current sector allocation and major holdings exposure (with up-to-date figures)~(0.12)}}, leaf]
      [{\parbox[c]{5.9cm}{Historical performance coverage with risk context (data-supported)~(0.15)}}, leaf]
      [{\parbox[c]{5.9cm}{Benefits coverage: diversification, market coverage, and practical investor advantages~(0.12)}}, leaf]
      [{\parbox[c]{5.9cm}{Risks and limitations coverage (structural + investor-facing)~(0.14)}}, leaf]
      [{\parbox[c]{5.9cm}{ETF implementation breadth: product selection and ownership mechanics~(0.14)}}, leaf]
      [{\parbox[c]{5.9cm}{Portfolio use as a core holding (fit, complements, and coverage gaps)~(0.04)}}, leaf]
    ]
    [{\parbox[c]{2.7cm}{\centering\textbf{Insight}\\\tiny weight: 0.23}}, branch
      [{\parbox[c]{5.9cm}{Interpretation of index construction \& composition implications (concentration, exposures, what ``World'' really means)~(0.18)}}, leaf]
      [{\parbox[c]{5.9cm}{Depth and specificity of risk/limitation mechanism analysis (not generic risk lists)~(0.18)}}, leaf]
      [{\parbox[c]{5.9cm}{Data-to-conclusion reasoning quality using historical performance (inference without overreach)~(0.16)}}, leaf]
      [{\parbox[c]{5.9cm}{Analytical trade-off framework for MSCI World as a core holding (when it fits, when it doesn't)~(0.16)}}, leaf]
      [{\parbox[c]{5.9cm}{ETF implementation insight: translating product features into investor outcomes~(0.14)}}, leaf]
      [{\parbox[c]{5.9cm}{Portfolio construction synthesis (complementary exposures and coherent allocation logic)~(0.10)}}, leaf]
      [{\parbox[c]{5.9cm}{Forward-looking and regime-sensitivity awareness (uncertainty, scenario thinking, behavioral pitfalls)~(0.08)}}, leaf]
    ]
    [{\parbox[c]{2.7cm}{\centering\textbf{Instruction}\\\textbf{Following}\\\tiny weight: 0.27}}, branch
      [{\parbox[c]{5.9cm}{Explains what the MSCI World Index represents and states key characteristics~(0.17)}}, leaf]
      [{\parbox[c]{5.9cm}{Describes main benefits, explicitly including diversification and market coverage~(0.17)}}, leaf]
      [{\parbox[c]{5.9cm}{Identifies primary risks and limitations investors should be aware of~(0.17)}}, leaf]
      [{\parbox[c]{5.9cm}{Provides practical considerations specifically for investing via MSCI World ETFs~(0.17)}}, leaf]
      [{\parbox[c]{5.9cm}{Uses current, explicit data for composition, sector allocation, and historical performance (with time reference)~(0.24)}}, leaf]
      [{\parbox[c]{5.9cm}{Maintains scope and audience fit: individual-investor guide focused on MSCI World as a core holding~(0.08)}}, leaf]
    ]
    [{\parbox[c]{2.7cm}{\centering\textbf{Readability}\\\tiny weight: 0.16}}, branch
      [{\parbox[c]{5.9cm}{Guide-like overall structure and section hierarchy~(0.20)}}, leaf]
      [{\parbox[c]{5.9cm}{Language clarity, concision, and mechanical correctness~(0.17)}}, leaf]
      [{\parbox[c]{5.9cm}{Terminology control and just-in-time definitions~(0.08)}}, leaf]
      [{\parbox[c]{5.9cm}{Logical flow, signposting, and transitions~(0.10)}}, leaf]
      [{\parbox[c]{5.9cm}{Information packaging and density management~(0.12)}}, leaf]
      [{\parbox[c]{5.9cm}{Table/chart design and labeling for composition, sectors, and performance~(0.12)}}, leaf]
      [{\parbox[c]{5.9cm}{Numerical context and interpretability of metrics~(0.10)}}, leaf]
      [{\parbox[c]{5.9cm}{Formatting, layout, and visual scannability~(0.07)}}, leaf]
      [{\parbox[c]{5.9cm}{Consistency and navigability (naming, time windows, cross-references)~(0.04)}}, leaf]
    ]
  ]
\end{forest}%
}
\caption{Rubric tree for the MSCI World Index open-ended task.
         Compared to Figure~\ref{fig:ex-sub-rubric-apple}, weight shifts toward
         \textit{comprehensiveness} and \textit{instruction following}, reflecting the expository
         nature of the question.}
\label{fig:ex-sub-rubric-msci}
\end{figure}
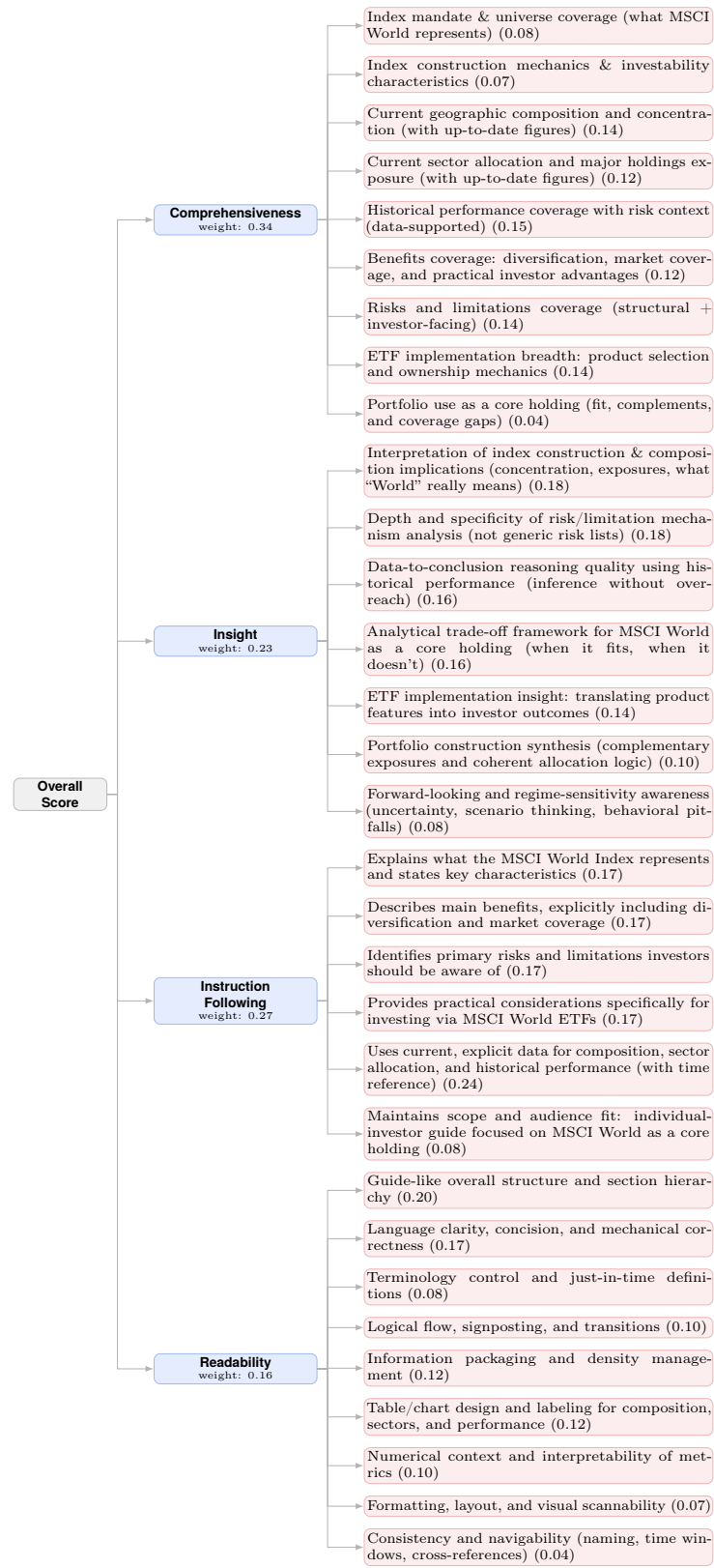

\paragraph{Rubric Tree.}
Figure~\ref{fig:ex-sub-rubric-msci} shows the rubric tree for this how-to-guide style question.
Note the contrast with the Apple example (Figure~\ref{fig:ex-sub-rubric-apple}): \textit{comprehensiveness}
becomes the dominant dimension (0.34 vs.\ 0.30), \textit{instruction following} rises (0.27 vs.\ 0.24)
because the four prompted sections must each be covered, and \textit{insight} drops (0.23 vs.\ 0.34)
since the task is primarily expository rather than argumentative.

\newpage

\section{Related Work}
\paragraph{Deep Research Agents.}
Recent advances extend LLMs with web search capabilities for complex knowledge-intensive tasks in domains such as finance and science~\citep{hu2025finsearchcomp,tang2026airesearcher}. 
These systems are often referred to as \textit{deep research agents}.
Motivated by this promise, many deep research systems have been developed, including OpenAI DeepResearch~\citep{openai2025deepresearch}, Gemini DeepResearch~\citep{gemini_dr}, Claude Research~\citep{anthropic2025claude_research}, and Kimi Researcher~\citep{moonshot2025kimi_researcher}.
Despite their effectiveness, these systems are based on proprietary models, and their data and training recipes are not publicly available.
Recent open-weight efforts have begun to improve this situation.
For example, Tongyi DeepResearch~\citep{team2025tongyi, li2025websailor, wu2025webdancer,wu2025resum} proposes a complete training framework that combines mid-training, supervised fine-tuning, and reinforcement learning to train open-weight LLMs~\citep{yang2025qwen3} as agents.
Other efforts, such as DR Tulu~\citep{shao2025drtulureinforcementlearning}, RedSearcher~\citep{chu2026redsearcher}, and OpenResearcher~\citep{li2026openresearcherfullyopenpipeline}, further explore scaling agent training data through a range of {automatic} data generation strategies, such as model distillation and graph-based task synthesis.
Despite recent progress, open-source deep research agents still lag behind proprietary systems in overall performance.
In this work, we propose an effective and comprehensive training recipe that combines data synthesis, context management, and staged post-training to further narrow this gap.

\paragraph{Deep Research Tasks.}
Traditional multi-hop question answering benchmarks, such as HotpotQA~\citep{yang2018hotpotqa} and 2WikiMultiHopQA~\citep{ho2020constructing}, measure a narrower set of capabilities, requiring models to seek evidence and perform multi-hop reasoning, usually to produce short answers.
Recent deep research benchmarks broaden these requirements toward long-horizon agentic search.
BrowseComp~\citep{wei2025browsecomp} and BrowseComp-Plus~\citep{chen2025BrowseCompPlus} emphasize search depth, focusing on hard-to-find, entangled facts that require persistent web navigation, whereas WideSearch~\citep{wong2025widesearchbenchmarkingagenticbroad} emphasizes search width through a large-scale collection of individually verifiable atomic facts.
Humanity's Last Exam (HLE;~\citet{phan2025humanity}) stresses knowledge-intensive reasoning over highly challenging expert-level queries, benefiting from external search.
GAIA~\citep{mialon2023gaia} evaluates real-world search-and-reasoning tasks, where agents must combine web browsing, tool use, and multi-step inference to produce concise factual answers.
Beyond fact seeking, Mind2Web 2~\citep{gou2025mind2web2} evaluates long-horizon web tasks with verifiable constraints and source attribution, requiring agents to return answers grounded in supporting links.
Open-ended benchmarks, such as DeepResearch Bench~\citep{du2025deepresearch} and LiveResearchBench~\citep{wang2026liveresearchbenchlivebenchmarkusercentric}, evaluate a further capability: synthesizing information from diverse sources into coherent, well-structured, and reader-friendly long-form reports.
Together, these benchmarks suggest that deep research agents must integrate multiple capabilities, including \capOne, \capTwo, and \capThree. 
Despite the complementary nature of these three capabilities, existing benchmarks and agent systems evaluate or support them only in isolation, and no prior work addresses them jointly within a unified framework.
\ours\ recipe aims to train a \textit{general-purpose} deep research agent to fulfill the gap.

\end{document}